%% file: root.tex
\documentclass[conference]{IEEEtran}

\usepackage{caption}
\usepackage{graphics}
\usepackage{epsfig}
\usepackage{mathptmx}
\usepackage{times}
\usepackage{amsmath}
\usepackage{amssymb}
\usepackage{blindtext}
\usepackage{multicol}
\usepackage{hyperref}
\usepackage{subcaption}
\usepackage{bbm}
\usepackage{microtype}
\usepackage{graphicx}
\usepackage{booktabs}
\usepackage[utf8]{inputenc}
\usepackage[T1]{fontenc}
\usepackage[noadjust]{cite}
\usepackage{wrapfig,lipsum}
\usepackage{caption}
\usepackage{nicefrac}
\usepackage{microtype}
\usepackage{epstopdf}
\usepackage[dvipsnames]{xcolor}
\usepackage{txfonts}
\usepackage[boxed]{algorithm2e}
\usepackage{pifont}
\usepackage{soul}
\usepackage[switch]{lineno}
\usepackage[numbers]{natbib}

\newcommand{\cmark}{\ding{51}}%
\newcommand{\xmark}{\ding{55}}%

\input{math_commands}

\title{\LARGE \bf
\textit{SPIRIT}: Perceptive Shared Autonomy for Robust \\ Robotic Manipulation under Deep Learning Uncertainty
}

\let\oldtwocolumn\twocolumn
\renewcommand\twocolumn[1][]{%
    \oldtwocolumn[{#1}{
    \begin{center}
           \includegraphics[width=1.0\textwidth]{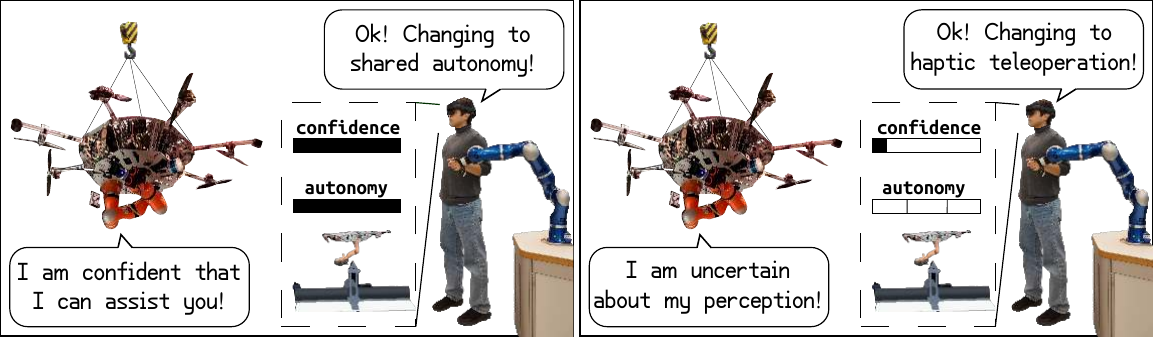}
           \captionof{figure}{Nowadays, Deep Learning (DL) is being widely used for robotic perception. However, DL cannot always be $100\%$ correct and can thus fail unexpectedly. A reliable robot needs to cope with such uncertainties in DL. We propose a shared autonomy concept that captures uncertainty in DL-based perception and transitions the level of autonomy for robustness.}
           \label{fig:fig1}
           \vspace*{-2mm}
        \end{center}
    }]
} 

\author{
\authorblockN{
Jongseok~Lee\authorrefmark{1}\authorrefmark{4}, Ribin~Balachandran\authorrefmark{1}, Harsimran~Singh\authorrefmark{1}, Jianxiang~Feng\authorrefmark{1}, Hrishik~Mishra\authorrefmark{1}, \\ Marco~De~Stefano\authorrefmark{1}, Rudolph~Triebel\authorrefmark{1}\authorrefmark{2}, Alin~Albu-Schäffer\authorrefmark{1}\authorrefmark{3} and Konstantin~Kondak\authorrefmark{1}
}
\authorblockA{\authorrefmark{1}DLR, \authorrefmark{2}KIT \authorrefmark{3}TU Munich \authorrefmark{4}Correspondence to {\tt\small jongseok.lee@dlr.de}}
}

\begin{document}

\maketitle

\input{chapters/abstract}

\IEEEpeerreviewmaketitle

\input{chapters/introduction}
\input{chapters/relatedwork}
\input{chapters/methods}
\input{chapters/results}
\input{chapters/conclusion}

\appendices
\input{chapters/appendix}

\bibliographystyle{plainnat}
\bibliography{bibliography}

\end{document}

%% file: math_commands.tex
\usepackage{amsmath,amsfonts,bm}

\def\eqref#1{equation~\ref{#1}}

\def\1{\bm{1}}

\def\vf{{\bm{f}}}

\def\vk{{\bm{k}}}

\def\vm{{\bm{m}}}

\def\vp{{\bm{p}}}
\def\vq{{\bm{q}}}

\def\vt{{\bm{t}}}

\def\vw{{\bm{w}}}
\def\vx{{\bm{x}}}
\def\vy{{\bm{y}}}

\def\mF{{\bm{F}}}

\def\mI{{\bm{I}}}
\def\mJ{{\bm{J}}}
\def\mK{{\bm{K}}}

\def\mP{{\bm{P}}}
\def\mQ{{\bm{Q}}}
\def\mR{{\bm{R}}}

\def\mT{{\bm{T}}}

\def\mX{{\bm{X}}}

\DeclareMathAlphabet{\mathsfit}{\encodingdefault}{\sfdefault}{m}{sl}
\SetMathAlphabet{\mathsfit}{bold}{\encodingdefault}{\sfdefault}{bx}{n}

\makeatletter
\DeclareRobustCommand\onedot{\futurelet\@let@token\@onedot}
\def\@onedot{\ifx\@let@token.\else.\null\fi\xspace}

\makeatother

\newcommand{\approptoinn}[2]{\mathrel{\vcenter{
  \offinterlineskip\halign{\hfil$##$\cr
    #1\propto\cr\noalign{\kern2pt}#1\sim\cr\noalign{\kern-2pt}}}}}

%% file: chapters/abstract.tex
\begin{abstract}


Deep learning (DL) has enabled impressive advances in robotic perception, yet its limited robustness and lack of interpretability hinder reliable deployment in safety-critical applications. We propose a concept termed perceptive shared autonomy, in which uncertainty estimates from DL-based perception are used to regulate the level of autonomy. Specifically, when the robot’s perception is confident, semi-autonomous manipulation is enabled to improve performance; when uncertainty increases, control transitions to haptic teleoperation for maintaining robustness. In this way, high-performing but uninterpretable DL methods can be integrated safely into robotic systems. A key technical enabler is an uncertainty-aware DL-based point cloud registration approach based on the so-called Neural Tangent Kernels (NTK). We evaluate perceptive shared autonomy on challenging aerial manipulation tasks through a user study of 15 participants and realization of mock-up industrial scenarios, demonstrating reliable robotic manipulation despite failures in DL-based perception. The resulting system, named SPIRIT, improves both manipulation performance and system reliability. SPIRIT was selected as a finalist of a major industrial innovation award.

\end{abstract}

%% file: chapters/introduction.tex
\section{Introduction}
\label{sec:introduction}

In the fall of 1998, the interactive tour guide robot Minerva \citep{thrun2000probabilistic} was successfully deployed in the Smithsonian Museum. During its two weeks of operation, the robot traversed over 44 km while interacting with thousands of visitors. The key technical enabler was Minerva’s probabilistic algorithms. For perception, Minerva relied on probabilistic localization that explicitly estimated uncertainty in the robot’s state \citep{fox2000probabilistic}. This information was then used to generate actions such that the robot could accomplish its mission despite uncertainty in sensing and motion \citep{roy1999coastal}. In essence, Minerva showed how probabilistic approaches can enable robust operation in complex environments, where methods that ignored uncertainty failed to scale. Inspired by systems such as Minerva, probabilistic robotics led to subsequent progress over decades \citep{dellaert2012factor, Fox-RSS-19, Agha2021NeBulaQF}.

In contrast to the probabilistic foundations that enabled early deployments, modern robotics increasingly relies on deep learning (DL) for perception and decision-making \citep{billard2025roadmap}. Indeed, DL methods have enabled remarkable advances. Yet, their internal mechanisms remain difficult to interpret and their predictions can fail unexpectedly \citep{sunderhauf2018limits, gunning2019xai}, especially when test conditions are under-presented in training data \citep{feng2023topology, schnaus2023learning, lee2020estimating}. As a result, integrating DL methods into high-stakes robotic applications remains challenging, not because such models are ineffective, but because their uncertainty is rarely accounted for at the system level. In this paper, we demonstrate that uncertainty-aware system design, rather than model perfection, is what enables the robust operation of robots -- a lesson learned once in robotics through Minerva and revisited here for DL.

\begin{table*}
\centering
\caption{The main novelty of the proposed system within the current state-of-the-art in probabilistic shared autonomy.}
\begin{tabular}{lcccc}
\toprule
\multicolumn{1}{c}{} & \multicolumn{1}{c}{\textbf{Perception uncertainty}} & \multicolumn{1}{c}{\textbf{Equipped with}} & \multicolumn{1}{c}{\textbf{Equipped with}} & \multicolumn{1}{c}{\textbf{Validation on}} \\
\multicolumn{1}{c}{} & \multicolumn{1}{c}{\textbf{from Deep Learning}} & \multicolumn{1}{c}{\textbf{haptic feedback}} & \multicolumn{1}{c}{\textbf{3D visual feedback}} & \multicolumn{1}{c}{\textbf{floating-base systems}} \\
\midrule
\textbf{Learning human intent} (i.e., \citep{hara2023uncertainty, Zhao-RSS-24, yow2023shared, zurek2021situational})  & \xmark  & \cmark & \xmark  & \xmark \\
\textbf{Adaptive authority allocation}  (i.e., \citep{palmieri2024perception, balachandran2020adaptive, saeidi2018incorporating})  & \xmark & \cmark & \xmark & \xmark \\
\textbf{Probabilistic virtual fixtures} (i.e., \citep{muhlbauer2024probabilistic, aarno2005adaptive, raiola2018co})  & \xmark  & \cmark & \xmark  & \xmark \\
\textbf{Interactive and imitation learning} (i.e., \citep{hoquethriftydagger, zurek2021situational})  & \xmark  & \xmark & \xmark  & \xmark \\
\midrule
\textbf{The proposed system: SPIRIT}  & \cmark   & \cmark & \cmark & \cmark\\
\bottomrule
\end{tabular}
\vspace*{-5mm}
\label{table1:relatedworks}
\end{table*}

This paper proposes a concept called perceptive shared autonomy for robust robotic manipulation under DL uncertainty (see Fig. \ref{fig:fig1}). With our concept, even if the used DL methods provide erroneous predictions when perceiving the environments, the robot is still able to accomplish the given tasks. We achieve this by allowing the robot's users to actively transition between semi-autonomous manipulation and haptic teleoperation (``shared autonomy") \citep{selvaggio2021autonomy}. Concretely, we model the uncertainty of the robot's DL-based perception \citep{lee2022trust}. If the uncertainties are low, we transition to semi-autonomous manipulation for better performance. When the uncertainties are high, the user can transition to haptic teleoperation, opting for robustness. Here, special emphasis is placed on designing interactive features for the users. Thus, a human-robot interface is designed to intuitively provide the users with haptic and visual feedback about the robot's current notion of perceptual uncertainty.

We evaluate the proposed concept on aerial manipulation tasks, where robots manipulate objects while airborne, making this domain a suitable testbed for safety-critical applications due to the high operational risk. The developed system, named SPIRIT, is the main focus of this paper.  In particular, our technical development is centered on SPIRIT's DL-based perception, as uncertainty quantification in DL remains an open research challenge \citep{gawlikowski2023survey}. To address this, we propose a partitioned approach to point cloud registration, in which a digital twin of the environments is divided into local regimes. Within each regime, DL models are trained to perform prediction. Uncertainty is quantified using a Gaussian Process (GP) with a DL-based kernel called the Neural Tangent Kernel (NTK) \citep{lee2022trust}. Empirically, we first provide ablation studies to motivate the design of SPIRIT's perception system. Robustness gains and overall system performance are evaluated within a user study with 15 participants and demonstration of industrial scenarios. SPIRIT's prototype was deployed for a major industrial exhibition performing aerial manipulation demonstrations in public over five consecutive days. Later, SPIRIT was selected as a finalist of an innovation award.

\textbf{Contributions and major claims.} Our main contribution is the development of SPIRIT, demonstrating robust robotic manipulation under DL uncertainty. We coin the term perceptive shared autonomy because we modulate the level of autonomy based on uncertainties in DL-based perception (Section~\ref{sec:sharedautonomy}). Our human-robot interface with haptic and 3D visual feedback that communicates the robot's uncertainty is another novelty of this paper (Section~\ref{sec:sharedautonomy:users}). To realize SPIRIT, we propose a partitioned approach that learns to register point clouds with GP-based uncertainty estimation (Section~\ref{sec:perception}). We claim that our partitioned approach and GP-based uncertainty estimation enable SPIRIT's development (Section \ref{sec:results:ablation}). SPIRIT also empirically improves the robustness of the overall system by achieving the robotic manipulation tasks, even when DL-based perception has unexpectedly failed (Section~\ref{sec:results:userstudy} and \ref{sec:results:industrialscenario}). The video accompanying this paper illustrates SPIRIT’s system components and its behavior under DL uncertainty while performing challenging aerial manipulation tasks.

Our project website can be found at \url{https://sites.google.com/view/robotspirit}. SPIRIT was originally developed for an industrial exhibition. The website disseminates those efforts.

%% file: chapters/relatedwork.tex
\section{Related Work}
\label{sec:relatedwork}

\begin{figure*}
    \centering
    \includegraphics[width=1.0\linewidth]{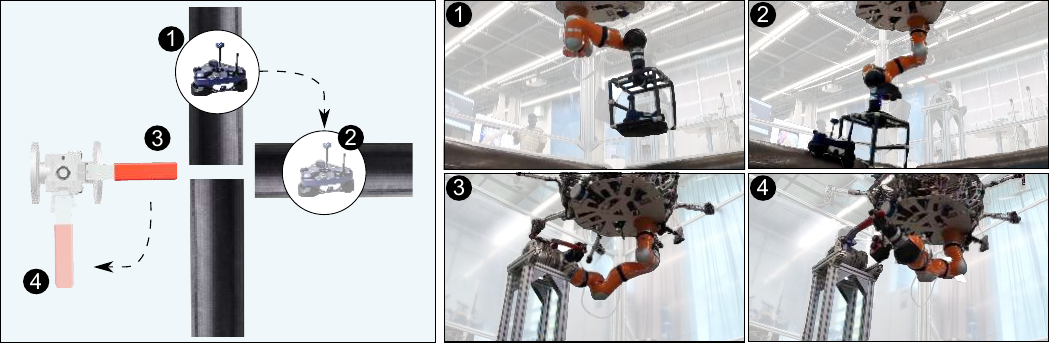}
    \caption{Amongst many potential applications, we examine two use cases: (a) extending the mobility of an inspection robot via pick and place, and (b) performing manipulation of industrial flange valves. The demonstration of these use cases along with the task sequence (1-4) is illustrated. Left: overview of the tasks. Right: robot performing the tasks from the left image.}
    \label{fig:concept}
    \vspace*{-7mm}
\end{figure*}


Our contribution lies in shared autonomy for robotic manipulation, building on probabilistic perception with DL. Tab. \ref{table1:relatedworks} summarizes the novelty: (a) leveraging uncertainty in DL-based perception, and (b) enabling user interaction with this uncertainty through haptic and XR. We validate these ideas on a moving (floating) base system such as aerial manipulation.

\textbf{Shared autonomy.} Complete autonomy in aerial manipulation is still a difficult goal. An alternative is to rely on human supervision \cite{lee2020visual, coelho2021hierarchical, lee2024}. One classical way to interface a human operator with a robot is through shared autonomy, which combines the benefits of autonomy and direct teleoperation \cite{selvaggio2021autonomy}. A common example is assistive driving, where onboard autonomy supports human drivers, reducing cognitive effort. For shared autonomy, determining the level of autonomy for the human remains an open research question \cite{selvaggio2021autonomy}.

To address this challenge, probabilistic methods have been widely investigated. Notable examples are on learning human intent with DL. Recent works \cite{hara2023uncertainty, Zhao-RSS-24, yow2023shared, zurek2021situational} use uncertainty estimates in the context of DL-based human intent prediction for improving assistive robot autonomy. Probabilistic methods are also applied for virtual fixtures (VFs) -- artificial forces that guide the human operator to achieve the given tasks \cite{muhlbauer2024probabilistic, aarno2005adaptive, raiola2018co}. Adaptive control \cite{palmieri2024perception, balachandran2020adaptive, saeidi2018incorporating} and interactive learning with uncertainty \cite{hoquethriftydagger, zurek2021situational} have been also investigated.

\textbf{Aerial manipulation.} Equipping aerial systems with manipulation capabilities is an active area of research \cite{ollero2021past}. In this work, we use aerial manipulation as a testbed for evaluating shared autonomy. Unlike many fixed-base manipulation systems that can operate autonomously, aerial manipulation remains difficult to automate reliably in general, and human-in-the-loop paradigms are therefore being investigated \citep{lee2025human, lee2024, perozo2024teleoperation}. Our platform is based on a cable-suspended aerial manipulator \cite{kong2024suspended, sarkisov2019development, miyazaki2019long}, where a robotic arm is mounted on a self-stabilizing suspended platform. However, the focus of this paper is not on aerial manipulation hardware or control, but on shared autonomy and DL-based perception. Detailed descriptions of system design, control, and field experimentation of such platforms can be found in existing works \cite{coelho2021hierarchical, sarkisov2019development, lee2024}.

\textbf{Uncertainty in deep learning.} This work employs and adapts a probabilistic method \cite{lee2022trust} that provides the so-called sampling-free uncertainty estimates for DL, i.e.,  the method does not require sampling \cite{gal2016dropout} or model ensembles \cite{lakshminarayanan2017simple}, thereby more suited for deployments on a real robot. We refer to the recent surveys for more information \cite{gawlikowski2023survey}. We note that our goal is not to advance state-of-the-art uncertainty quantification. Instead, we attempt to demonstrate that an uncertainty-aware autonomy can improve overall system reliability against potential errors in DL methods.

%% file: chapters/methods.tex
\section{Perceptive Shared Autonomy under Uncertainty in DL}
\label{sec:sharedautonomy}

Among others, we ground the development of SPIRIT in an application within the oil and gas industry, where over 825 refineries operate globally. These facilities require regular maintenance, with some refineries having 40,000 km of pipes and 50,000 inspection routines \citep{ollero2018aeroarms}. To automate this, robotic crawlers are being developed, but many pipes are in hard-to-reach areas, limiting mobility. Another challenge is human safety, particularly with industrial valves, which can fail and release hazardous substances. In 2018, a valve malfunction at Husky Energy led to an explosion and fatal injuries.

With SPIRIT, can we extend the mobility of robotic inspection crawlers and operate industrial valves without human presence in proximity? Motivated by this problem, we designed a testbed and a demonstration scenario as depicted in Fig.~\ref{fig:concept}. In our demonstration scenario, first, the robot grasps a cage that hosts an inspection crawler robot. Second, the robot picks up the cage and places the cage onto another pipe. Due to the gap between three pipes in the setup, the crawler robot that uses magnetic wheels cannot traverse from one pipe to another. Thus, the first use case extends the mobility of the crawlers for inspection tasks. Next, the robot attempts to close an industrial valve. For this, the robot grasps the handle of the flange valve, and the robot rotates the valve. Then, the robot releases the grasp, avoiding any collision. We note that this demonstration scenario evaluates SPIRIT's design with practical relevance. We do not claim to solve an industrial problem in this work.

\subsection{The Design of SPIRIT - Shared Autonomy Concept}

In aerial manipulation, teleoperation is being actively investigated \citep{lee2025human, lee2024, perozo2024teleoperation}. Teleoperation is generally motivated by the ability of human operators to maintain reliable task execution when autonomy is limited, reducing the likelihood of mission failure and improving resilience of the overall system.

Yet, teleoperation demands high physical and mental effort from human operators \citep{selvaggio2021autonomy}. To alleviate this, shared autonomy has been explored, where both the robot and the human operator collaborate to complete a task. When the robot's autonomy struggles or fails, the human operator steps in to assist, thereby improving efficiency and task success rates. Among others, we use the so-called mixed-initiative shared autonomy, in which, the weights called \textit{authority allocation factor} $\alpha$ decide how much the task is shared by the human and the robot's autonomy. In a broader sense, mixed initiative can be described by:
\begin{equation}
\dot{\bm{s}}(t) = f(\bm{s}(t), \bm{a}(t)) \ \ \text{with} \ \ \bm{a}(t) = \alpha \bm{a}_h(t) + (1-\alpha) \bm{a}_a(t),
\end{equation}
where $\bm{s}$ is the state, $\bm{a}$ is the control input, and $f(\cdot)$ describes the system dynamics. The control input is composed of a weighted addition of human input $\bm{a}_h$ and robot autonomy $\bm{a}_a$. Within this framework, our key idea is to decide the authority allocation factor $\alpha$ based on uncertainty in DL-based perception. 

We use two torque-controlled manipulators: one mounted on a suspended platform (Fig. \ref{fig:fig1}) and another as a haptic device for force-feedback teleoperation. The platform stabilizes with propellers, while its manipulator performs the tasks. The control inputs are Cartesian wrenches $\mF \in \mathbb{R}^{6}$ which are converted into joint torques $\bm{\tau} = \mJ^T \mF$. Here, $\mJ$ is the manipulator Jacobian. Now, using the haptic device, the human operator controls the robotic manipulator on the platform. This means, at time t, the human sends position information $\bm{s}_{\text{h}}$ (velocity analogously). Due to the communication time delay T, the robot receives $\bar{\bm{v}}_{\text{r}}(t) = \bm{v}_{\text{h}}(t-T)$ or equivalently $\bar{\bm{s}}_{\text{r}}(t) = \bm{s}_{\text{h}}(t-T)$. If controller gains are denoted with K, and $\bm{v}_{\text{r}}, \bm{s}_{\text{r}}$ are measured quantities, the commanded wrench $\bm{a}_h(t) = \mF_{\text{h}}$ from the operator is:
\begin{equation}
    \bm{a}_h(t) = \mF_{\text{h}} = K_{d, \text{r}}(\bar{\bm{v}}_{\text{r}}(t) - \bm{v}_{\text{r}}(t)) + K_{p, \text{r}}(\bar{\bm{s}}_{\text{r}}(t) - \bm{s}_{\text{r}}(t)).
\end{equation}
At the same time, the robot's onboard autonomy assists the human operator in achieving the desired task. Imagine the visual perception of the robot's target pose $\bm{s}_{\text{a}}$ (or $\bm{v}_{\text{a}})$). Then,
\begin{equation}
    \bm{a}_a(t) = \mF_{\text{a}} = K_{d, \text{a}}(\bm{v}_{\text{a}}(t) - \bm{v}_{\text{r}}(t)) + K_{p, \text{a}}(\bm{s}_{\text{a}}(t) - \bm{s}_{\text{r}}(t)),
\end{equation}
is the action from the autonomy. For providing guidance to the human operator, we generate the wrench command $\mF=\alpha \mF_{\text{h}} + (1-\alpha) \mF_{\text{a}}$ to the robot. In return, the robot sends these forces back to the operator for haptic. The guiding forces $\mF_{\text{a}}$ act as VFs, resulting in semi-autonomous manipulation.

Within this framework, a key concept in SPIRIT is how the authority allocation factor $\alpha$ is chosen. To recap, VFs are created by the robot's onboard perception system, for which DL is being widely adopted. We assume that the DL-based perception system returns Gaussian distributions with a mean vector and a covariance matrix. Section \ref{sec:perception} provides an ample description of our perception system. The mean vector is used to create VFs directly. The covariance provides a measure of uncertainty. Using these uncertainty estimates, we decide the level of authority with a metric $\left\| \mathbf{\Sigma} \right\|$ and a threshold $\beta$:
\begin{equation}
\label{eq:shared_autonomy}
    \alpha = 1 \quad \text{if} \quad \left\| \mathbf{\Sigma} \right\| > \beta \quad \text{else} \quad \frac{1}{2}.
\end{equation}
When $\alpha = \frac{1}{2}$, VFs are turned on for high performance, enabling semi-autonomous manipulation. Otherwise, when $\alpha = 1$, VF is turned off, opting for robustness through haptic teleoperation. In this way, SPIRIT safely leverages VFs from DL-based perception by turning them off when uncertainty is high. Closely following \textit{Balachandran et.al.} \citep{balachandran2020adaptive}, we use the trace as our metric $\left\| \mathbf{\Sigma} \right\| = \text{tr}(\mathbf{\Sigma})$, which captures the total variance. Furthermore, the threshold $\beta$ is chosen from a validation data set that contains nominal working conditions, i.e., we take an upper bound of the uncertainty metric. While these design choices provide sufficient means for SPIRIT, we note that more elaborated control strategies \citep{balachandran2020adaptive, hoquethriftydagger} can also be leveraged. 

\begin{figure}
    \centering
    \includegraphics[width=0.995\linewidth]{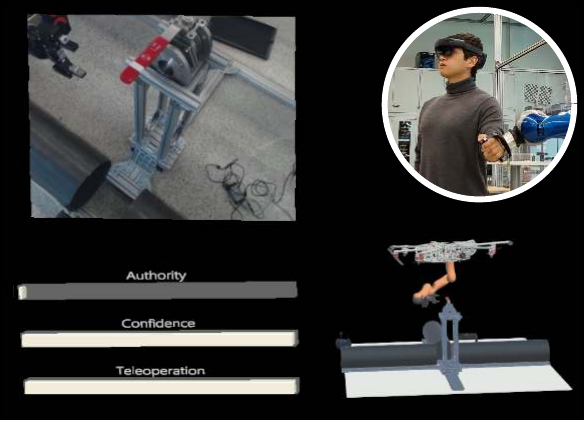}
    \caption{User interface of SPIRIT that communicates robot's notion of uncertainty to a human operator. In addition to haptic, we use XR for 2D and 3D visualization, as well as telemetry.}
    \label{fig:interface} 
    \vspace*{-5mm}
\end{figure}

\subsection{User Interfaces of SPIRIT}
\label{sec:sharedautonomy:users}

In our design of SPIRIT, the user interacts with the robot and its uncertainty through haptic and visual feedback. A torque-controlled manipulator serves as the haptic device, providing an extended workspace, while we integrate a Microsoft HoloLens 2 for 2D and 3D visual feedback: live camera streams (2D) and robot/environment state visualization (3D). The operator can use hand gestures to zoom, adjust views, and interact with the 3D environment during aerial manipulation. Audio feedback is also integrated in the system. Importantly, telemetry such as authority level ($\alpha$) and uncertainty or confidence ($\left\| \mathbf{\Sigma} \right\|$) is displayed, and a foot pedal allows manual authority adjustment, ensuring flexible remote operation. A previous study \citep{balachandran2020adaptive} suggested that an option for manual human intervention can be useful, for which both authority and confidence should be displayed. Fig.~\ref{fig:interface} depict the proposed user interface visually.

\begin{figure*}
    \centering
    \includegraphics[width=1.0\linewidth]{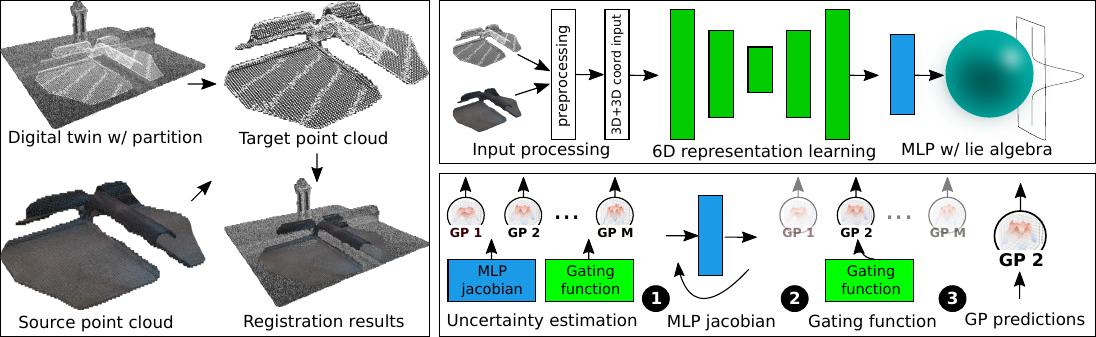}
    \caption{Left: the proposed partitioned approach to point cloud registration. Matching source point cloud to the target point cloud (partitioned from the digital twin) is easier than the use of the entire point cloud from the digital twin of environments. Target point cloud is captured based on robot's position in the digital twin when performing manipulation tasks. Right: the proposed registration architecture. Lie algebra (a 6D vector) is inferred from neural networks (top). Then, Gaussian Processes (GPs) are utilized for uncertainty estimation (bottom). Uncertainty estimates are inferred over predictions in lie algebra.}
    \label{fig:uncertainty}
    \vspace*{-5mm}
\end{figure*}

\section{Uncertainty-Aware Perception in SPIRIT}
\label{sec:perception}

Estimating target pose of objects and its uncertainty are the main outcomes of SPIRIT's perception system. A key requirement is reliable uncertainty estimates, i.e., if pose errors are large, uncertainty should be high, and vice versa. Consequently, our goal is not to advance state-of-the-art perception in accuracy, but to design a pipeline whose uncertainty is reliable in order to realize and test perceptive shared autonomy. Thus, grounded in our industrial scenario, we develop a perception system, which consists of a point cloud registration network (Section~\ref{sec:perception:a}) and an uncertainty estimation algorithm (Section~\ref{sec:perception:b}). An overview of our approach is further depicted in Fig.~\ref{fig:uncertainty}. More background information and algorithmic details are in appendix.

\subsection{Learning to Register Point Clouds with Digital Twins} 
\label{sec:perception:a}

\textbf{Problem.} Assume that the robot has a 3D sensor to obtain the point clouds $\mP =  \begin{bmatrix}
\vp_1, & \cdots, & \vp_P \\
\end{bmatrix} \in \mathbb{R}^{3 \times P}$. The sensor is installed without occlusions to the environments. In our industrial scenario in Fig.~\ref{fig:concept}, there are known objects such as a cage and an inspection robot. For them, we can use fiducial markers like April-tags \citep{wang2016apriltag} due to their accuracy and robustness. We also have objects that belong to the environments, such as industrial pipes and valves. For them, we cannot use April-tags since industrial sites are typically not equipped with markers. 

Digital twins of industrial sites are becoming a reality \citep{xiangdong2020asset}. Similar to high-definition maps for autonomous driving, 3D models of industrial environments with pipelines and valves can support robotic perception. A key benefit is the simplification of the perception pipeline. Prior work relied on complex systems involving object detection and 3D reconstruction \citep{lee2020visual, lee2024}. Under the above assumptions, perception reduces to point cloud registration: current sensor measurements are aligned with a reference point cloud from the digital twin. The resulting alignment pose localizes the robot in the environment. Since objects such as pipelines and valves are static, robot localization directly enables 6D object pose estimation via the digital twin.

However, point cloud registration in real-world is challenging. For example, finding the points from one point cloud that correspond to points in another point cloud, often referred to as the correspondence estimation step, might be difficult if two point clouds significantly differ. The digital twin contains a 3D model of the entire environments, while the robot's sensors may only provide local measurements. As these local measurements constitute only a small portion of the entire environments, finding correspondences is not trivial. In Fig.~\ref{fig:uncertainty} (left), for an intuition, (a) source point cloud (local measurements) and (b) digital twin (entire environment) differ in shape and density.

\textbf{Approach.} Hence, we propose a partitioning approach. Our key idea is to partition the 3D model of the entire environments, so that the given registration problem is made easier. Intuitively, as illustrated in Fig.~\ref{fig:uncertainty} (left), we partition the digital twin into (a) target point cloud $\mQ = \begin{bmatrix}
\vq_1, & \cdots, & \vq_Q \\
\end{bmatrix} \in \mathbb{R}^{3 \times Q}$, which is easier to align with (b) source point cloud $\mP$. Since we know the alignment between digital twin and target point cloud (simply from partitioning), registering the target and source point cloud can align digital twin and source point cloud. This results in localization of the robot within the entire environment. We note that the target point cloud can be obtained from the digital twin, considering the robot's position for manipulability. This is justified because when performing manipulation tasks, mobile robots often regulate its base fixed around one position. 

A digital twin also enables the collection of training data for DL-based point cloud registration. We design a pipeline that aligns the point clouds $\mP$ and $\mQ$, as depicted in Fig.~\ref{fig:uncertainty} (right). Firstly, we process the point clouds $\mP$ and $\mQ$ by estimating the features $\mathcal{F}_\mP = (\vf_{\mP_1},...,\vf_{\mP_P})$ and $\mathcal{F}_\mQ = (\vf_{\mQ_1},...,\vf_{\mQ_Q})$. Then, we use the nearest neighbor to generate the correspondences $\mathcal{M} = \left\{ (i, \text{argmin}_j \left\| \vf_{\mP_i} - \vf_{\mQ_j}  \right\| )\right\}$ for $i=1,...P$ given $j=1,...,Q$. We utilize these correspondences of features as a sparse input $\vx$ to a neural network $f_\mathbf{\theta}$, which consists of a U-net-like 6D convolutional network \citep{choy2020deep} and a multi-layer perceptron (MLP):
\begin{equation}
\label{eq:network}
    \vy = f_\mathbf{\theta}(\vx) \ \text{with} \ \vx=\begin{bmatrix}
\vf_{\mP_1} & \cdots & \vf_{\mP_{|\mathcal{M}|}} \\
\vf_{\mQ_{J_1}} & \cdots & \vf_{\mQ_{J_{|\mathcal{M}|}}} \\
\end{bmatrix}.
\end{equation}
Here, the indices $J$ define the correspondences $\vf_{\mP_i} \leftrightarrow \vf_{\mQ_{J_i}}$.
The network outputs a 6D representation called Lie algebra $\vy$ that can be mapped into transformation matrices \citep{barfoot2024state}, aligning $\mP$ and $\mQ$. Lie algebra is in Euclidean space as opposed to transformation matrices, allowing us to use Gaussians for uncertainty. We can optionally use the iterative closest point (ICP) to refine the obtained results further. The robot's onboard SLAM can also be used to track the environments and provide estimates of the object pose at a faster rate \citep{lee2020visual}. For training, we use a mean squared error as loss function. For a zero-shot Sim-to-Real transfer, we use a photorealistic synthesizer \citep{denninger2023blenderproc2} that generates training data by emulating 3D sensor measurements.

\subsection{Uncertainty Estimation with Gaussian Processes}
\label{sec:perception:b}

\textbf{Problem.} Next, we describe how we obtain uncertainty estimates. In the literature, uncertainty is commonly categorized into epistemic and aleatoric types \citep{gawlikowski2023survey}. Epistemic uncertainty reflects incomplete knowledge captured by the model, whereas aleatoric uncertainty arises from inherent randomness in the data-generating process, for example due to noise in training annotations in supervised learning. We quantify both types of uncertainty. A key challenge here is achieving computational efficiency while maintaining high-fidelity uncertainty estimates. 

\textbf{Approach.} We estimate uncertainty of our registration pipeline based on Gaussian Processes (GPs) -- probabilistic model that defines a distribution over possible functions and often referred as a gold standard in probabilistic machine learning \citep{rasmussen2003gaussian}. The pipeline is depicted in Fig.~\ref{fig:uncertainty} (right). The final output of the pipeline is a predictive distribution with a Gaussian parameterized by mean and covariance. 

The first step is to estimate the mean, for which we use the network to regress Lie algebra from a point cloud pair: $\vy = f_\mathbf{\theta}(\vx)$ (Eq.~\ref{eq:network}). Next, we estimate uncertainty over Lie algebra with a covariance matrix. The covariance is obtained from a variant of GP called mixtures of GP experts (MoE-GP). As depicted in Fig.~\ref{fig:uncertainty} (right bottom), MoE-GP is composed of an ensemble of GPs (or experts) and a gating function. During training, the gating function divides the input space into different regimes, assigning each regime to a specific GP expert. At test time, the gating function assigns each input to the corresponding expert (GP) to make the covariance estimation:
\begin{equation*}
\begin{aligned}
\begin{matrix}
\mathbf{\Sigma} = \vk_{**} - \vk^{T}_{*}(\mK+\sigma_n \mI)^{-1}\vk_{*} + \sigma_n \mI \ w/ \ k(\cdot,\cdot) = \mJ_f(\cdot)^T\mJ_f(\cdot).
\end{matrix}
\end{aligned}
\end{equation*}
The kernel $k(\cdot,\cdot)$ defines the covariances, capturing similarity between pairs of data points. Here, $\vk_{*} = k(\vx^*, \mX)$ is the row vector of covariances between new test input $\vx^*$ and training data $\mX$ (column vector), and $\mK=k(\mX,\mX)$ is the covariance matrix from the training data. Likewise, $\vk_{**}=k(\vx^*,\vx^*)$. Our kernel, NTK, is defined as an inner product of neural network Jacobians $\mJ_f$ with respect to its parameters. This kernel allows us to use the GP formulation for neural networks \citep{lee2022trust}. We note that, in GPs, the term $\sigma_n$ captures aleatoric uncertainty. We capture model uncertainty over MLP only, which is more efficient. Hence we compute NTK for only MLP, as in Fig.~\ref{fig:uncertainty}.

This formulation, originally from \citet{lee2022trust}, is well suited to SPIRIT. First, the covariance estimation is analytical, making it sampling-free. This means our pipeline does not introduce significant run-time overhead due to covariance computations. Second, GPs are known to provide reliable uncertainty estimates \citep{rasmussen2003gaussian}, which is crucial for SPIRIT because we use estimated covariance for modulating the level of autonomy (see Eq.~\ref{eq:shared_autonomy}). Finally, MoE-GPs reduce the computational complexity of GPs by partitioning the data into different regimes. For our use case, GPs are computationally tractable.

In addition, we introduce specific adaptations from \citet{lee2022trust} tailored to our use case, including (a) input-dependent non-stationary aleatoric uncertainty, (b) a gating function aligned with the digital twin’s partitioning, and (c) efficient implementations of the GP formulation. Further implementation details are provided in the appendix for interested readers.

%% file: chapters/results.tex
\section{Results}
\label{sec:results}

This work develops SPIRIT to show that uncertainty-aware system design can enable more robust robotic manipulation. Through ablation studies, we motivate our design choices and consequently support the claim that SPIRIT development is facilitated by our DL-based perception and uncertainty pipeline. Our claim on system-level robustness against DL uncertainty is demonstrated through a user study and completion of industrial scenarios. Overall, these experimental results support our contributions on the successful development of SPIRIT.

\subsection{Ablation Studies on SPIRIT's Perception} 
\label{sec:results:ablation}

First, we provide ablation studies on SPIRIT's perception module. Specifically, we analyze our design choices in the proposed approach to partition the point cloud registration problem with uncertainty estimates. For this ablation study, we plug in our digital twin to Blenderproc software in order to generate photo-realistic images with point clouds. An advantage of using such synthetic data is the availability of ground truth for noise-free evaluation. Furthermore, the data can also be used to train DL methods, where depth-only approaches suffer less from the sim-to-real gap; e.g., we adapt zero-shot transfer in all our real-world experiments. Emulating our demonstration scenario in Fig. \ref{fig:concept}, we pick four regimes of interest (or partitions equivalently). For each regime, we sample 1200 viewpoints from a sphere with a variance of 0.2 m, from which we split the train and test set with a ratio of 0.8. Regarding the evaluation metrics, we choose run-time, mean squared error (MSE) over rotations and translations, and test negative log likelihood (NLL) which is often used measure of uncertainty \citep{gawlikowski2023survey}.

\begin{figure}
\centering
\begin{subfigure}{.495\textwidth}
    \centering
    \includegraphics[width=.99\linewidth]{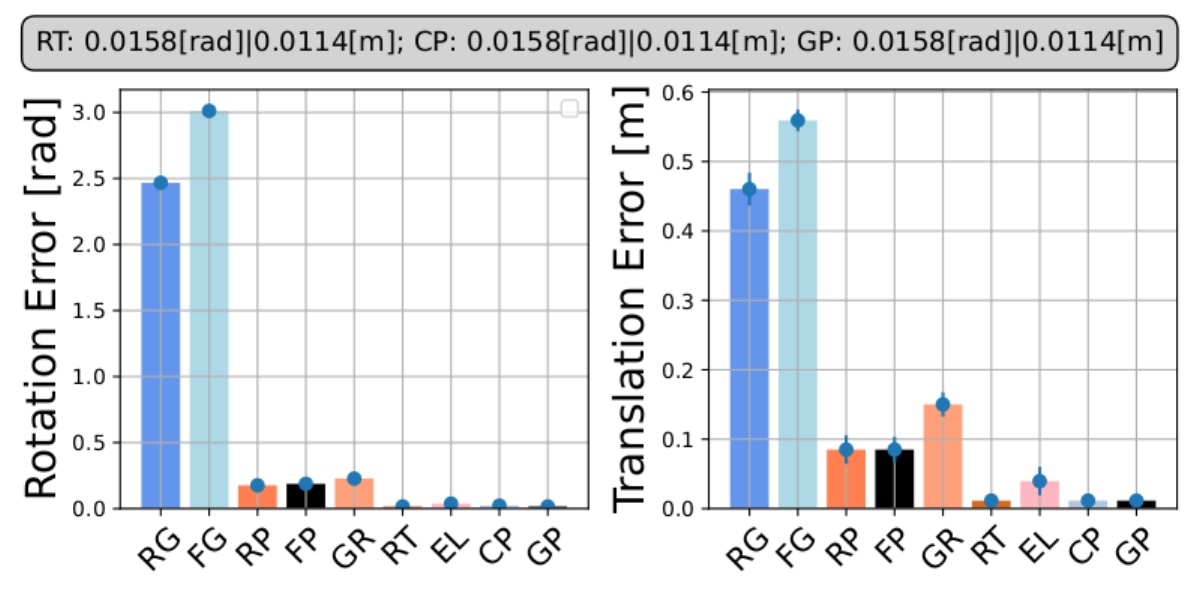}  
\end{subfigure} \hfill
\begin{subfigure}{.495\textwidth}
    \centering
    \includegraphics[width=.99\linewidth]{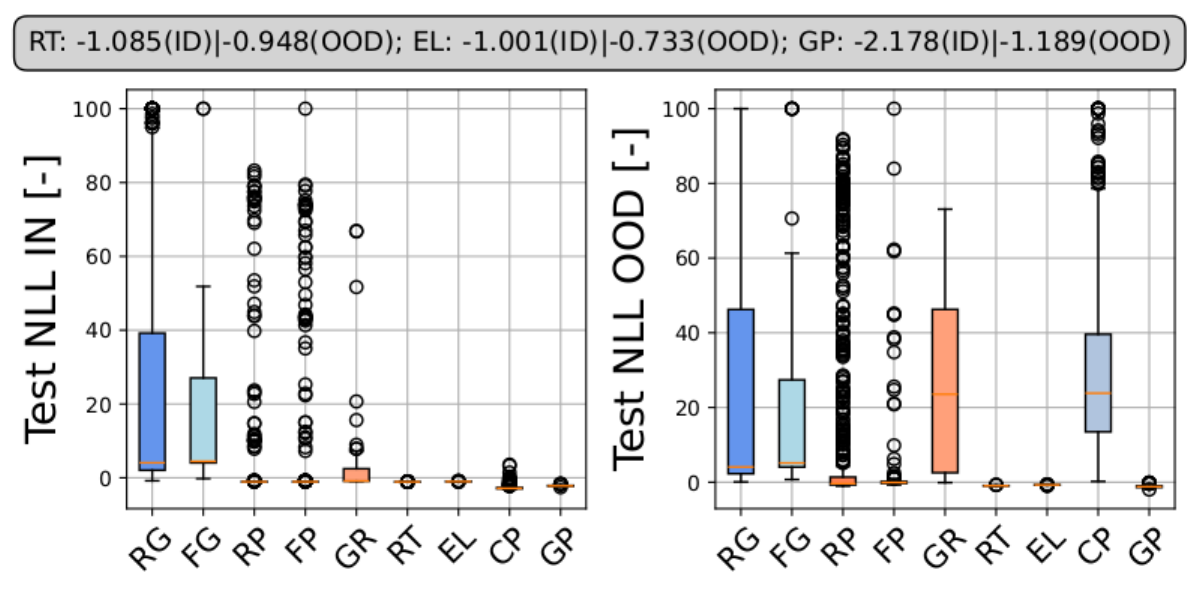}  
\end{subfigure}
\caption{Results of ablation studies. Lower the better. Values from the best three baselines are shown in legend. Lower rotation and translation errors are observed for the baselines relying on partitioning. GP (our approach) produced more reliable uncertainty estimates for nominal conditions (ID) and failure cases (OOD), indicated by lower NLL.}
\label{fig:results:ablation}
\vspace*{-5mm}
\end{figure}

In total, we chose eight baselines for ablation studies. First, we implement Open3D's RANSAC-based global registration (RG) and fast global registration (FG). Then, next two baselines are their combinations with the proposed partitioned approach with both RANSAC (RP) and fast global registration (FP). In this way, we can examine the influence of our partitioned approach. Then, while utilizing the partitioned approach, we compare deep global registration with the weighted procrustes (GR \citep{choy2020deep}) and the proposed architecture (RT). This comparison can motivate the design of our underlying DL architecture. For these deterministic methods, we modeled homoscedastic aleatoric uncertainties. Finally, with RT, we ablate different uncertainty estimation methods. We note that the goal is to examine the performance in our specific application scenario. Therefore, we compare the proposed approach with NTK-based GPs (GP) against two recent baselines, namely evidential learning (EL) \citep{amini2020deep} and conformal predictions (CP) \citep{fontana2023conformal}. Unlike MC-dropout and Deep Ensembles \citep{gawlikowski2023survey}, these baselines are sampling-free methods and are run-time efficient \citep{amini2020deep,fontana2023conformal}. These methods have gained popularity in recent years.

\begin{figure*}
\centering
\begin{subfigure}{.25\textwidth}
    \centering
    \includegraphics[width=1.0\linewidth]{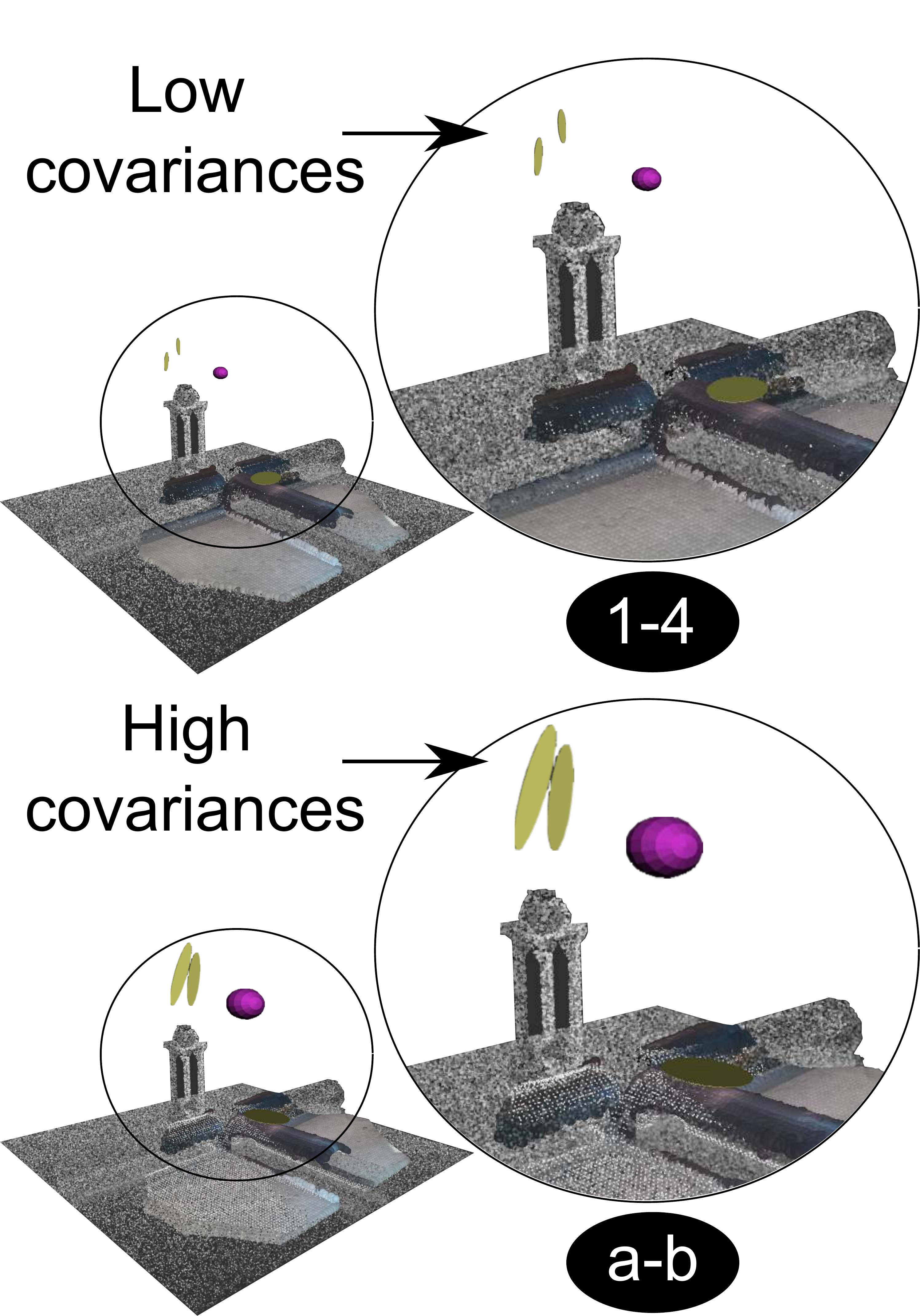}  
\end{subfigure}
\begin{subfigure}{.735\textwidth}
    \centering
    \includegraphics[width=.98\linewidth]{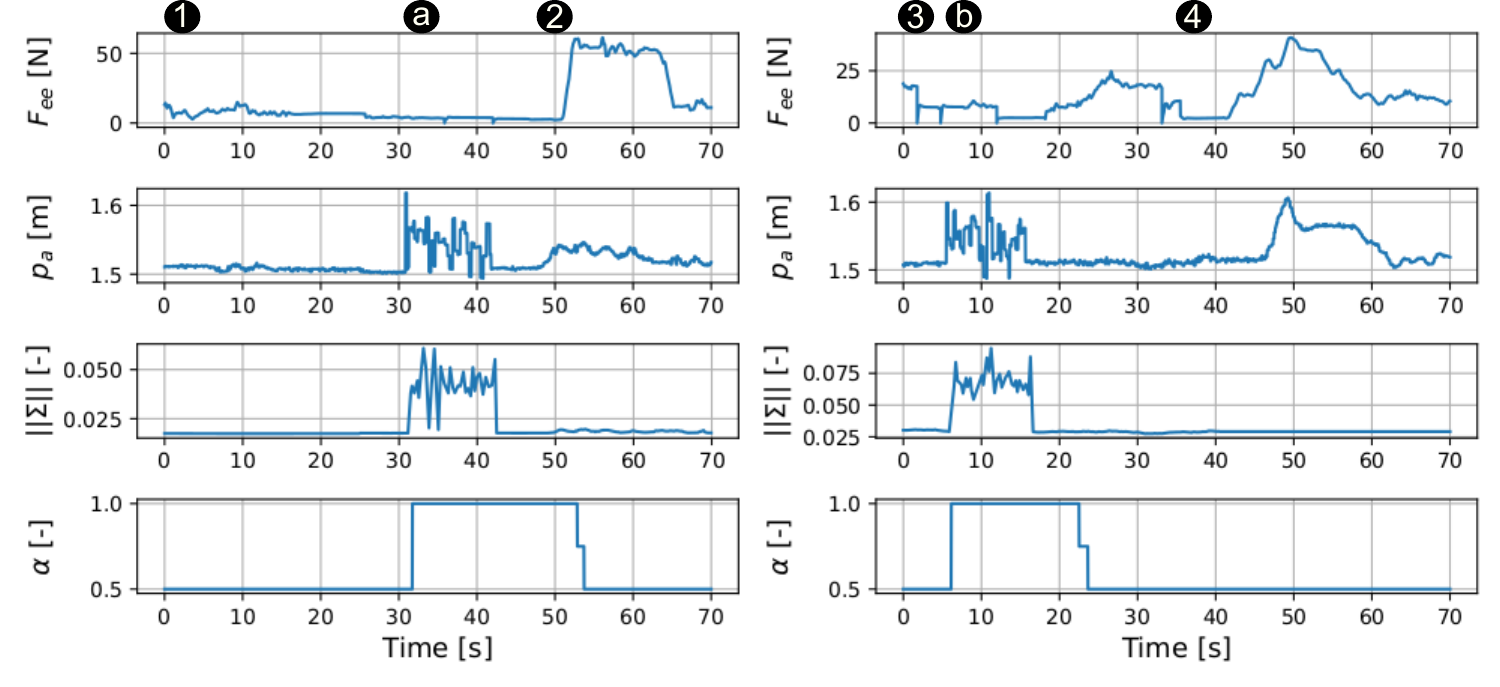}  
\end{subfigure}
\caption{Results from the demonstration of industrial scenarios. Left: the point cloud registration is visualized with covariances. Blue ellipse represent the covariance over the translations while yellow ellipse depict the covariance over rotations for each axis. Middle: pick-and-place of heavy objects; (1) grasping the object, (a) failures in DL-based perception over 10 s, and (2) pick-and-place with over 50 N end-effector forces. Right: forceful operation of flange valve; (3) grasping of the valve bar, (b) failures in DL-based perception while grasping, and (4) closing of the valve with about 50N end-effector forces (norm).}
\label{fig:results:industrial}
\vspace*{-5mm}
\end{figure*}

We not only examine the baselines on nominal conditions (ID) but also simulate their failure cases (OOD). In point cloud registration problems, sensor noise, symmetry, and lack of sufficient overlap (or occlusions) between two point clouds are typical failure cases. For SPIRIT, symmetry and lack of overlap are mitigated by the partitioning of the point cloud. Therefore, we simulate OOD by randomly perturbing a measured point cloud. To simulate challenging conditions, we randomly applied Gaussian noise with mean up to 0.25 m and covariance up to 0.5 m. These values were chosen to stress-test the system beyond typical sensor specifications, exposing potential failure modes. Simulating these failure cases is crucial because SPIRIT needs its perception to produce higher uncertainty when registration is likely to fail and lower uncertainty under nominal conditions, enabling reliable decision-making for shared autonomy.

The results are depicted in Fig. \ref{fig:results:ablation}. First, partitioned variants consistently improve accuracy compared to non-partitioned counterparts, and learning-based baselines achieve lower errors overall. These observations support the use of partitioned approach combined with DL for point cloud registration. Second, in terms of uncertainty estimation, probabilistic baselines achieve lower test NLL. However, EL degrades accuracy, potentially due to its non-standard loss formulation and architectural modifications. In our experiments, conformal prediction does not improve uncertainty estimates. In terms of run-time, sampling-free methods, namely EL, CP and GP induced minimum overhead ($\pm0.1$ FPS) to RT. Overall, GP achieves the best trade-off between test NLL, accuracy, and runtime. Consequently, within our evaluation, our design satisfies the requirements of SPIRIT’s perception system — providing reliable uncertainty for failure detection while maintaining accuracy and run-time efficiency.

\subsection{User Studies on SPIRIT's Shared Autonomy} 
\label{sec:results:userstudy}

We conduct a user study in order to evaluate SPIRIT in assisting a human operator for the given manipulation tasks, while coping with uncertainty in DL. T testing sample of $15$ consisting of 12 male and three female is considered. On average, the subjects had an age of 30.93 years (SD$=$3.76 with a range 26-39). All the subjects had little to no experience in teleoperation and XR. An informed consent form was signed for the study. We instructed the subjects to complete the valve grasping task (see Fig. \ref{fig:concept}) remotely. That is, no direct view of the robot was provided while executing the tasks.

We divided the study into two sets. First, we compare three operation modes: bilateral haptic teleoperation with XR (SPIRITv1), unilateral teleoperation without XR (vanilla-teleop), and the proposed system (SPIRITv2) to assess shared autonomy benefits. The mode vanilla-teleop is included to evaluate our user interface, where the subject was only provided with a video stream on a computer monitor. The order was randomized. Second, we evaluate the robustness against failures in the DL-based perception. Two modes are examined, namely the proposed system and a system without uncertainty (vanilla-VF). Here, we manually fail the perception by perturbing a point cloud with a random Gaussian noise from our ablation study. While SPIRIT can change its authority, vanilla-VF would result in a complete failure. In order to avoid breaking the hardware, we emulated the failure by introducing a sinusoidal disturbance to the VF with a 10 Hz frequency and 0.1 m amplitude. The order was counterbalanced across the subjects.

\begin{table}
	\centering
	\caption{Results from the user study. Baselines (Vanila-teleop and VF) are compared against variants of SPIRIT.}
	\label{table:userstudy}
		\begin{tabular}[ht]{lccc}
            \toprule
            \textbf{Set 1} & \textbf{Vanila-teleop} & \textbf{SPIRITv1} & \textbf{SPIRITv2} \\
            \midrule
                \textbf{Success rate [$\%$]} $\uparrow$ & 86.66$\pm$0.000 & 93.33$\pm$0.000 & \textbf{100.0$\pm$0.000} \\
                \textbf{Time [$s$]} $\downarrow$ & 160.46$\pm$144.1 & 149.1$\pm$112.3 & \textbf{61.86$\pm$46.03} \\
                \textbf{Forces [$N$]} $\downarrow$ & 11.66$\pm$1.325 & \textbf{10.89$\pm$0.867} & 11.72$\pm$1.414 \\
                \textbf{Torques [$Nm$]} $\downarrow$ & 6.674$\pm$0.748 & \textbf{6.351$\pm$0.381} & 7.442$\pm$0.810 \\
                \textbf{NASA TLX [-]} $\downarrow$ & 11.58$\pm$3.715 & 11.71$\pm$2.118 & \textbf{7.422$\pm$2.613} \\
                \textbf{SUS score [-]} $\uparrow$ & 51.00$\pm$25.73 & 52.66$\pm$19.41 & \textbf{71.33$\pm$16.77} \\
            \bottomrule
            \toprule
            \textbf{Set 2} & \textbf{Vanila-VF} & {\textbf{SPIRIT}} &  \multicolumn{1}{|c|}{\textbf{p} (set 1 / 2)} \\
            \midrule
                \textbf{Success rate [$\%$]} $\uparrow$ & 40.00$\pm$0.000 & \textbf{100.0$\pm$0.000} & \multicolumn{1}{|c|}{(N/A / N/A)} \\
                \textbf{Time [$s$]} $\downarrow$ & 335.0$\pm$202.2 & \textbf{78.33$\pm$72.77} & \multicolumn{1}{|c|}{(0.0189/0.0249)} \\
                \textbf{Forces [$N$]} $\downarrow$ & 13.70$\pm$1.018 & \textbf{10.86$\pm$1.586} & \multicolumn{1}{|c|}{(N.S/0.0010)} \\
                \textbf{Torques [$Nm$]} $\downarrow$ & 7.202$\pm$0.449 & \textbf{6.679$\pm$0.903} & \multicolumn{1}{|c|}{(0.0004/N.S)} \\
                \textbf{NASA TLX [-]} $\downarrow$ & 14.18$\pm$3.416 & \textbf{8.288$\pm$4.137} & \multicolumn{1}{|c|}{(0.0002/0.0003)} \\
                \textbf{SUS score [-]} $\uparrow$ & 35.83$\pm$20.92 & \textbf{65.50$\pm$24.10} & \multicolumn{1}{|c|}{(0.0245/0.0016)} \\
            \bottomrule
		\end{tabular}
	\vspace*{-5mm}
\end{table}%

Subjects were first introduced to SPIRIT’s concept, features, and tasks, then completed a demographic questionnaire. Initially, the subjects were trained to move the robot until they felt ready, after which the experiments began. For each set of experiments, subjects were administered the NASA Task Load Index (TLX) and System Usability Scale (SUS), which qualitatively measure the overall workload and usability. At the end, a situation awareness questionnaire was given using an eight-point scale. Completion time, success rate, mean forces, and torques were measured as quantitative measures. A task was assumed to have failed after 500 s or if it was decided to stop by the participant. Analysis of variance (ANOVA) and mean with standard deviation were calculated from recordings.

Results in Tab.~\ref{table:userstudy} show that set 1 highlights the benefits of shared autonomy (SPIRITv2) in success rate, completion time, workload (NASA TLX), and usability (SUS). Moreover, when compared to vanilla-teleop, the benefits of haptic and elaborated visual feedback (SPIRITv1) are observed in terms of success rate and completion time. Mean forces did not pass the statistical significance testing (N.S). The subjects used the least torque in SPIRITv1 and the most torque in SPIRITv2. Forces and torques reflect physical workload but may also indicate user confidence. Overall, the results from set 1 provide evidence that the concept of perceptive shared autonomy can improve the success rate and completion time of the users. The subjects also rated SPIRIT as more usable with less workload.

Set 2 results (Tab.~\ref{table:userstudy}) show SPIRIT outperforms vanilla-VF across all measures, with users achieving a 100$\%$ success rate despite DL-based perception failures. We note that the success rate of 40$\%$ with vanilla-VF may not be feasible in real failure cases of DL-based perception. Imagine the jump in the estimates of the target's rotation, which can damage the robot's hardware or result in a collision with the environment. In contrast, SPIRIT estimates DL uncertainty and turns false VFs off, allowing users to still complete tasks using teleoperation. These results suggest SPIRIT improves system reliability in employing DL-based perception across different users.

In the situational awareness questionnaires, participants reported moderate awareness of the robot’s position and actions ($5.062\pm1.611$), ability to predict its behavior ($5.187\pm1.515$), and perceived control ($6.125\pm1.204$). They generally found the haptic ($6.312\pm0.946$) and XR interfaces ($6.750\pm1.064$) helpful in understanding the robot's confidence and authority, which validates the design of our user interface.

\begin{figure}
    \centering
    \includegraphics[width=1.0\linewidth]{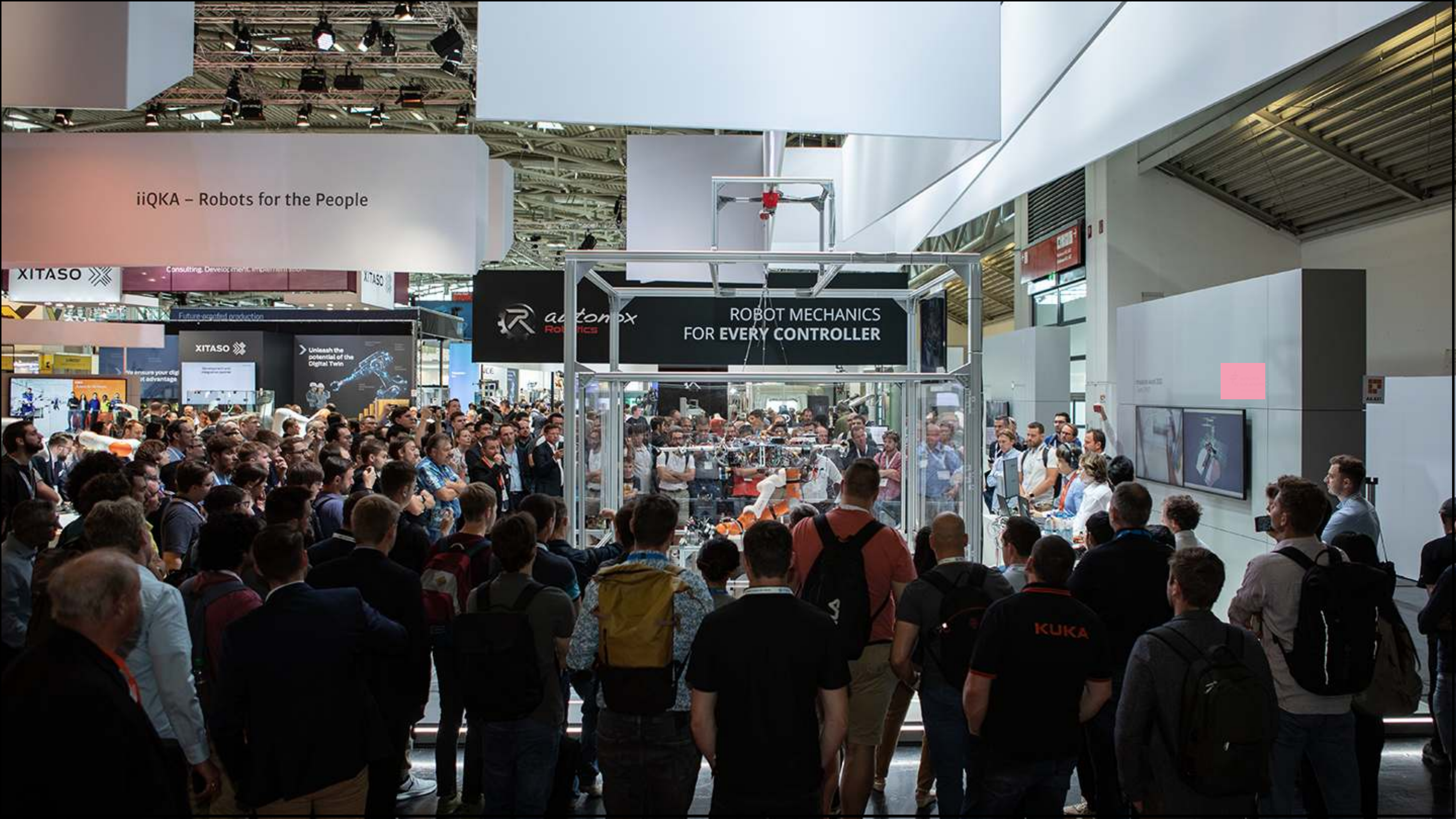}
    \caption{A photo from the exhibitions of SPIRIT at one of the world's largest industrial fair on robotics and automation.}
    \label{fig:panoramic}
    \vspace*{-5mm}
\end{figure}

\subsection{Completion of Industrial Scenarios with SPIRIT} 
\label{sec:results:industrialscenario}

We demonstrate SPIRIT in an industrial scenario introduced in Section~\ref{sec:sharedautonomy}, involving flange valve operations and pick-and-place of heavy objects. The scenario was designed for an industrial exhibition and implemented using a mock-up composed of real industrial pipes, inspection robots, and valves, as shown in Fig.~\ref{fig:panoramic}. To evaluate robustness under perceptual failures, we inject noise into the incoming point clouds while a human operator performs manipulation tasks in semi-autonomous mode with VFs. These perception failures are introduced unexpectedly, without informing the operator.

The results for both tasks are summarized in Fig.~\ref{fig:results:industrial}. The plots report the measured end-effector wrench $\mF_{ee}$ from the force–torque sensor, the perception output $\vp_a$ (visualized as geodesic distance of the 6D pose estimate for clarity of the plot), the trace of the estimated covariance $\lVert \mathbf{\Sigma} \rVert$ (see Eq.~\ref{eq:shared_autonomy}), and the authority level $\alpha$, where $\alpha = 1$ corresponds to pure teleoperation and $\alpha = \frac{1}{2}$ indicates that VFs are active. When a perception failure occurs, the quality of the DL-based perception output degrades significantly. Nevertheless, the tasks are completed successfully by reverting to teleoperation until perception accuracy is restored. The uncertainty criterion $\lVert \mathbf{\Sigma} \rVert$ reliably detects these failures and disables VFs accordingly. In both scenarios, the robot applies wrench forces of up to $50,\mathrm{N}$, demonstrating forceful aerial manipulation.

These results indicate that SPIRIT enables reliable robotic manipulation in the presence of uncertainty in DL-based perception. We achieve this by explicitly accounting for uncertainty at the system level. Within an industrially motivated scenario, SPIRIT improves system reliability by appropriately regulating autonomy during perceptual failures. In this way, in a similar vein to Minerva but within a modern context, SPIRIT shows the relevance of probabilistic approaches when employing DL methods in complex real-world settings. More implementation details and results are provided in the appendix.

%% file: chapters/conclusion.tex
\section{Conclusion}

We presented the concept of perceptive shared autonomy and a robotic system, SPIRIT, that realizes this concept by explicitly accounting for uncertainty in DL-based perception at the system level. SPIRIT combines an aerial manipulator, haptic and XR-based human interfaces, a shared autonomy with VFs, and a partitioned perception approach that provides actionable uncertainty estimates. Through extensive ablation studies against eight baselines, a 15-participant user study, and industrially motivated demonstrations involving forceful aerial manipulation tasks, we evaluated the proposed approach in a complex real-world application. The results show that SPIRIT maintains task execution under perception failures. Overall, this work demonstrates how uncertainty-aware system design enables robust deployment of DL-based perception in practice.

\textbf{Limitations and future work.} SPIRIT assumes that manipulation tasks remain feasible under teleoperation when autonomy is disabled, and cannot ensure task completion when teleoperation alone is insufficient. Moreover, while the proposed uncertainty estimation is validated in the controlled industrial scenarios, broader validation in real-world environments, including outdoor settings, will assess its functionalities under naturally occurring failures in DL-based perception.

\section*{Acknowledgements}

This work was supported by MOTIE AiSac under Grant RS-2024-00441872 and also partially by the German Federal Ministry of Research, Technology, and Space (BMFTR) under the Robotics Institute Germany (RIG). The authors also acknowledge Manuel Schnaus and Nari Song for their participation in our industrial exhibition campaign.

%% file: chapters/appendix.tex
\section{Appendix Overview}
This appendix is structured to provide background materials, algorithmic and implementation details, additional analyses, and system-level evidence that complement the main paper and support reproducibility. The contents are organized as follows:

\begin{itemize}
    \item \textbf{Preliminaries (Section~\ref{sec:preliminaries}).} Defines notation and summarizes the background needed for SPIRIT’s probabilistic formulation, including Gaussian Processes, Mixtures of Experts (hard partitioning), and Neural Tangent Kernels.

    \item \textbf{Perception and uncertainty (Section~\ref{sec:perception_uncertainty}).} Provides detailed algorithmic descriptions of the point cloud registration pipeline and uncertainty estimation, including the sparse 3D network architecture (Tab.~\ref{tab:resunet_mlprt_appendix}) and the mathematical derivations used for GP/NTK-based uncertainty.

    \item \textbf{System architecture (Section~\ref{sec:system_architecture}).} Describes the hardware and software stack, operator interface, and how uncertainty is integrated into the shared autonomy, its decision-making and feedback loop that utilizes uncertainty estimates.

    \item \textbf{User study details (Section~\ref{sec:user_study}).} Reports participant demographics and the devised situational awareness questionnaire items used in the qualitative evaluation.

    \item \textbf{Additional details and results (Section~\ref{sec:additional_details_results}).} Contains extended materials not included in the main paper:
    \begin{itemize}
        \item \textbf{Baseline implementations.} Unifies the evaluation protocol across classical, partitioned, learning-based, and uncertainty baselines (preprocessing, calibration, corruption tests, runtime measurement).
        \item \textbf{Runtime analysis.} Compares computational cost across all methods and shows that the proposed uncertainty estimation adds negligible overhead.
        \item \textbf{Industrial scenario.} Additional quantitative data from floating-base manipulation conditions.
        \item \textbf{Industrial exhibition.} Demonstrations at major robotics/automation fairs, including crawler deployment/retrieval and valve closing (Fig.~\ref{fig:demo1}--\ref{fig:demo2}).
    \end{itemize}
\end{itemize}

\noindent\textbf{Supplementary videos.}
\begin{itemize}
    \item \textbf{Video~1:} Main video with story of SPIRIT, containing concept explanations and main experimental results.
    \item \textbf{Video~2:} Crawler deployment and retrieval sequence (cage grasping, placement, pressing for release, and retrieval).
    \item \textbf{Video~3:} Valve closing demonstration (grasping and controlled turning under industrial constraints).
\end{itemize}

\section{Preliminaries}
\label{sec:preliminaries}

\noindent\textbf{Notation.}
We follow the notation of the main paper. Scalars are non-bold (e.g., $x\in\mathbb{R}$), vectors are bold lowercase (e.g., $\vx\in\mathbb{R}^d$), and matrices are bold uppercase (e.g., $\mK\in\mathbb{R}^{n\times n}$). A point cloud is denoted by $\mP=[\vp_1,\dots,\vp_{|\mP|}]\in\mathbb{R}^{3\times|\mP|}$ and a (partitioned) digital-twin reference by
$\mQ=[\vq_1,\dots,\vq_{|\mQ|}]\in\mathbb{R}^{3\times|\mQ|}$. Rigid transformations are $\mT\in SE(3)$ and their Lie algebra coordinates are $\vy=\log(\mT)\in\mathbb{R}^6$\footnote{Typically, $\boldsymbol{\xi}$ is used for Lie algebra coordinates. But, since we learn the coordinates and therefore they are outputs of neural network, we denote them as $\vy$ for notation simplicity.}.
Uncertainty is represented by a covariance $\boldsymbol{\Sigma}\in\mathbb{R}^{6\times 6}$. In shared autonomy, $\alpha\in[0,1]$ denotes the authority allocation factor and $\beta$ the uncertainty threshold.

\subsection{On Gaussian Processes}

A Gaussian Process (GP) \citep{rasmussen2003gaussian} is a collection of random variables such that any finite subset follows a joint Gaussian distribution. Equivalently, a GP defines a probability distribution over functions, fully specified by a mean function and a covariance function. GPs can be understood from two complementary perspectives: a generative model and a predictive inference framework. The generative view specifies a prior distribution over latent functions and a likelihood that relates these functions to observed data through noise. Conditioning this prior on observations yields the predictive distribution, which provides both point predictions and uncertainty estimates at new inputs. Together, these perspectives offer a coherent Bayesian treatment of regression that combines flexible function modeling (non-parametric) with principled uncertainty quantification.

\textbf{Generative model.} Let $\vx \in \mathbb{R}^d$ denote an input vector and $f(\vx)$ a latent real-valued function:
\begin{equation}
    f(\vx) \sim \mathcal{GP}\big(m(\vx), k(\vx, \vx')\big),
\end{equation}
where $m(\vx) = \mathbb{E}[f(\vx)]$ is the mean function, $k(\vx, \vx') = \mathrm{cov}(f(\vx), f(\vx'))$ is the covariance (kernel) function. We assume that observations are generated from this latent function with additive Gaussian noise $\varepsilon$ with $\sigma_n^2$ as the variance:
\begin{equation}
y = f(\vx) + \varepsilon
\qquad \text{with} \qquad
\varepsilon \sim \mathcal{N}(0, \sigma_n^2).
\end{equation}

Given a set of $n$ training inputs $\vx = \{\vx_1, \ldots, \vx_n\}$ and corresponding observations $\mathbf{y} = (y_1, \ldots, y_n)^\top$, the joint distribution of the observed outputs is:
\begin{equation}
\mathbf{y} \sim \mathcal{N}\big(\vm(\mX), \mK(\mX, \mX) + \sigma_n^2 \mI\big),
\end{equation}
where $\vm(\mX) = (m(\vx_1), \ldots, m(\vx_n))^\top$ and $\mK(\mX, \mX) \in \mathbb{R}^{n \times n}$ is the kernel matrix with entries $[\mK(\mX, \mX)]_{ij} = k(\vx_i, \vx_j)$.

The kernel function encodes prior assumptions about the latent function, such as smoothness, periodicity, or stationarity. GPs are non-parametric models: rather than assuming a fixed finite-dimensional parameterization, a GP defines a distribution over an infinite-dimensional function space. As more data are observed, the effective model complexity adapts automatically.

\textbf{Predictive inference.} Let $\vx_*$ denote a test input and $f_* = f(\vx_*)$ the corresponding latent function value. The predictive posterior distribution conditioned on the training data $(\vx, \mathbf{y})$ is Gaussian:
$p(f_* \mid \mX, \mathbf{y}, \vx_*) = \mathcal{N}(\mu_*, \sigma_*^2)$ with predictive mean:
\begin{equation}
\mu_* = \mK(\vx_*, \mX)\big[ \mK(\mX, \mX) + \sigma_n^2 \mI \big]^{-1} \mathbf{y},
\end{equation}
and predictive variance
\begin{equation}
\sigma_*^2 = k(\vx_*, \vx_*)
- \mK(\vx_*, \mX)\big[ \mK(\mX, \mX) + \sigma_n^2 \mI \big]^{-1} \mK(\mX, \vx_*),
\end{equation}
where $\mK(\vx_*, \mX) = \big(k(\vx_*, \vx_1), \ldots, k(\vx_*, \vx_n)\big)$ and $\mK(\mX, \vx_*) = \mK(\vx_*, \mX)^\top$. A key strength of GPs is the quantification of uncertainty. In regions of the input space where training data are sparse, the covariance vector $\mK(\vx_*, \mX)$ becomes small, causing the predictive variance $\sigma_*^2$ to revert toward the prior variance $k(\vx_*, \vx_*)$. This behavior provides an explicit measure of predictive confidence and signals when predictions are less reliable.

\subsection{On Mixtures of Experts}

The Mixture of Experts (MoE) model \citep{jacobs1991adaptive} approximates a global function by partitioning the input space into $M$ regions. The total predictive mapping is defined as:
\begin{equation}
    f(\vx) = \sum_{m=1}^{M} g_m(\vx) f_m(\vx),
\end{equation}
where $g_m(\vx)$ represents the responsibility assigned to the $m$-th expert (the gating function).

In \textit{hard partitioning}, the input space $\mathcal{X}$ is divided into disjoint sets $\{\mathcal{R}_m\}_{m=1}^M$ such that $\bigcup_m \mathcal{R}_m = \mathcal{X}$ and $\mathcal{R}_i \cap \mathcal{R}_j = \emptyset$ for $i \neq j$. A strict gating function is employed, typically taking the form of an indicator:
\begin{equation}
    g_m(\vx) = \mathbb{I}(\vx \in \mathcal{R}_m).
\end{equation}
This formulation ensures that for any input $\vx$, only the expert specialized for that specific region is activated.

The integration of GPs within a Mixture of Experts framework provides several advantages over monolithic GP models.

\textbf{Scalability and parallelization.}
The primary bottleneck of GPs is the $O(n^3)$ inversion of the covariance matrix. By employing hard partitioning of the input space into $M$ regions, the computational complexity is reduced to $O(M(n/M)^3)=O(n^3/M^2)$. This decomposition allows for distributed training where each expert $f_m$ performs inference on a local subset of the data $\mathcal{D}_m$ independently.

\textbf{Handling non-stationarity.}
Standard GP kernels are typically stationary, assuming uniform smoothness across the entire domain $\mathcal{X}$. In a GP-MoE, each expert $f_m$ can be assigned a unique set of hyperparameters $\boldsymbol{\theta}_m$:
\begin{equation}
    f_m(\vx) \sim \mathcal{GP}\big(0, k_{\boldsymbol{\theta}_m}(\vx, \vx')\big).
\end{equation}
This allows the model to capture non-stationary behavior by assigning experts with small lengthscales to high-frequency regions and experts with large lengthscales to smoother regions.

\subsection{On Neural Tangent Kernels}

The NTK provides a bridge between neural networks and kernel methods. For a neural network $f_{\boldsymbol{\theta}}(\vx)$ in the infinite-width limit, the training dynamics under gradient descent are governed by a kernel k($\cdot, \cdot$) with Jacobians $\mJ_f = \nabla_{\boldsymbol{\theta}} f_{\boldsymbol{\theta}}(\vx)$:
\begin{equation}
    k(\vx, \vx') = \mJ_f(\vx)^\top \mJ_f(\vx').
\end{equation}
These results provide a theoretical foundation for understanding sufficiently wide neural networks as kernel regressors. We leverage this connection for efficient uncertainty estimation via NTK-based approximations in this work.

\section{Perception and Uncertainty}
\label{sec:perception_uncertainty}

\begin{table*}[h]
\centering
\caption{Detailed architecture of the point cloud registration network. All convolutions are sparse Minkowski convolutions operating on 3D coordinates. BN denotes batch normalization.}
\label{tab:resunet_mlprt_appendix}
\begin{tabular}{l l c c c}
\hline
\textbf{Stage} & \textbf{Layer} & \textbf{Kernel} & \textbf{Stride} & \textbf{Channels} \\
\hline
Input & Sparse tensor input & -- & -- & $3$ \\
\hline
Encoder-1 
& Conv + BN + ResBlock & $3^3$ & 1 & $3 \rightarrow 32$ \\
Encoder-2 
& Conv + BN + ResBlock & $3^3$ & 2 & $32 \rightarrow 64$ \\
Encoder-3 
& Conv + BN + ResBlock & $3^3$ & 2 & $64 \rightarrow 128$ \\
\hline
Decoder-3 
& Transposed Conv + BN + ResBlock & $3^3$ & 2 & $128 \rightarrow 64$ \\
Decoder-2 
& Transposed Conv + BN + ResBlock & $3^3$ & 2 & $(64 + 64) \rightarrow 64$ \\
Decoder-1 
& Conv (skip fusion) & $1^3$ & 1 & $(32 + 64) \rightarrow 32$ \\
\hline
Output Head 
& Conv & $1^3$ & 1 & $32 \rightarrow C_{\text{out}}$ \\
\hline
Global Pooling 
& Max / Avg / Sum Pooling & -- & -- & -- \\
\hline
Feature Aggregation
& Concatenation of pooled features & -- & -- & $352$ \\
\hline
MLP Head
& Linear + ReLU & -- & -- & $352 \rightarrow 128$ \\
& Linear + ReLU & -- & -- & $128 \rightarrow 16$ \\
Rotation Head
& Linear & -- & -- & $16 \rightarrow 4$ \\
Translation Head
& Linear & -- & -- & $16 \rightarrow 3$ \\
\hline
\end{tabular}
\end{table*}

\subsection{Point Cloud Registration: Algorithmic Details}

We provide algorithmic details of our point cloud registration, including architecture, learning pipelines and data generation.

\textbf{Architecture.} The proposed network follows an encoder--decoder architecture based on sparse 3D convolutions implemented using MinkowskiEngine. The encoder progressively downsamples the input sparse point cloud while increasing feature dimensionality, whereas the decoder restores spatial resolution via transposed convolutions and skip connections. Each encoder stage consists of a $3 \times 3 \times 3$ sparse convolution followed by batch normalization and a residual block. Downsampling is performed using stride-2 convolutions. The decoder mirrors this structure using transposed convolutions with identical kernel sizes, and skip connections are implemented through feature concatenation. To capture global context at multiple resolutions, global pooling is applied after each encoder and decoder stage, as well as after the final output convolution. The pooled features are concatenated to form a global feature vector of dimension 352, which aggregates information across different spatial scales. The aggregated feature vector is processed by a multi-layer perceptron consisting of two fully connected layers with ReLU activations. The network outputs both rotation and translation parameters. The rotation head uses Lie algebra ($d_r = 3$) while translation is represented as a 3D vector.

\textbf{Training pipeline.} The network is trained as follows. Given a pair of input point clouds $(\mP, \mQ)$, we voxelize both point clouds using MinkowskiEngine’s sparse quantization with voxel size $\nu$. After quantization, each point cloud is represented as a sparse tensor with coordinates and a dummy feature vector of ones, which serves as input to the sparse convolutional backbone. FPFH features are extracted for each point cloud, and nearest-neighbor matching is performed to obtain an initial correspondence set $\{(i, j)\}$. For each point in $\mP$, the nearest neighbor in $\mQ$ is selected using GPU-based KNN search. For each correspondence $(i, j)$, an inlier feature vector is constructed which is used as input to the network. The network receives the correspondence features as a sparse tensor and predicts a rotation $\hat{\mR}$ and translation $\hat{\vt}$. Lie algebra is used for rotation.

When the rotation is represented using Lie algebra \citep{barfoot2024state}, the network outputs a vector $\hat{\boldsymbol{\omega}}\in\mathbb{R}^3$. This vector is interpreted as an axis-angle representation, where the rotation angle is $\|\hat{\boldsymbol{\omega}}\|$ and the rotation axis is $\hat{\boldsymbol{\omega}} / \|\hat{\boldsymbol{\omega}}\|$. The corresponding rotation matrix is obtained using the exponential map:
\begin{equation}
\hat{\mR} = \exp(\hat{\boldsymbol{\omega}}_\times),
\end{equation}
where $\hat{\boldsymbol{\omega}}_\times$ denotes the skew-symmetric matrix of $\hat{\boldsymbol{\omega}}$:
\begin{equation}
\hat{\boldsymbol{\omega}}_\times =
\begin{bmatrix}
0 & -\omega_3 & \omega_2 \\
\omega_3 & 0 & -\omega_1 \\
-\omega_2 & \omega_1 & 0
\end{bmatrix}.
\end{equation}
The estimated transformation is constructed as:
\begin{equation}
\hat{\mT} =
\begin{bmatrix}
\hat{\mR} & \hat{\vt} \\
\mathbf{0}^\top & 1
\end{bmatrix}.
\end{equation}

\begin{figure}
    \centering
    \includegraphics[width=0.495\textwidth]{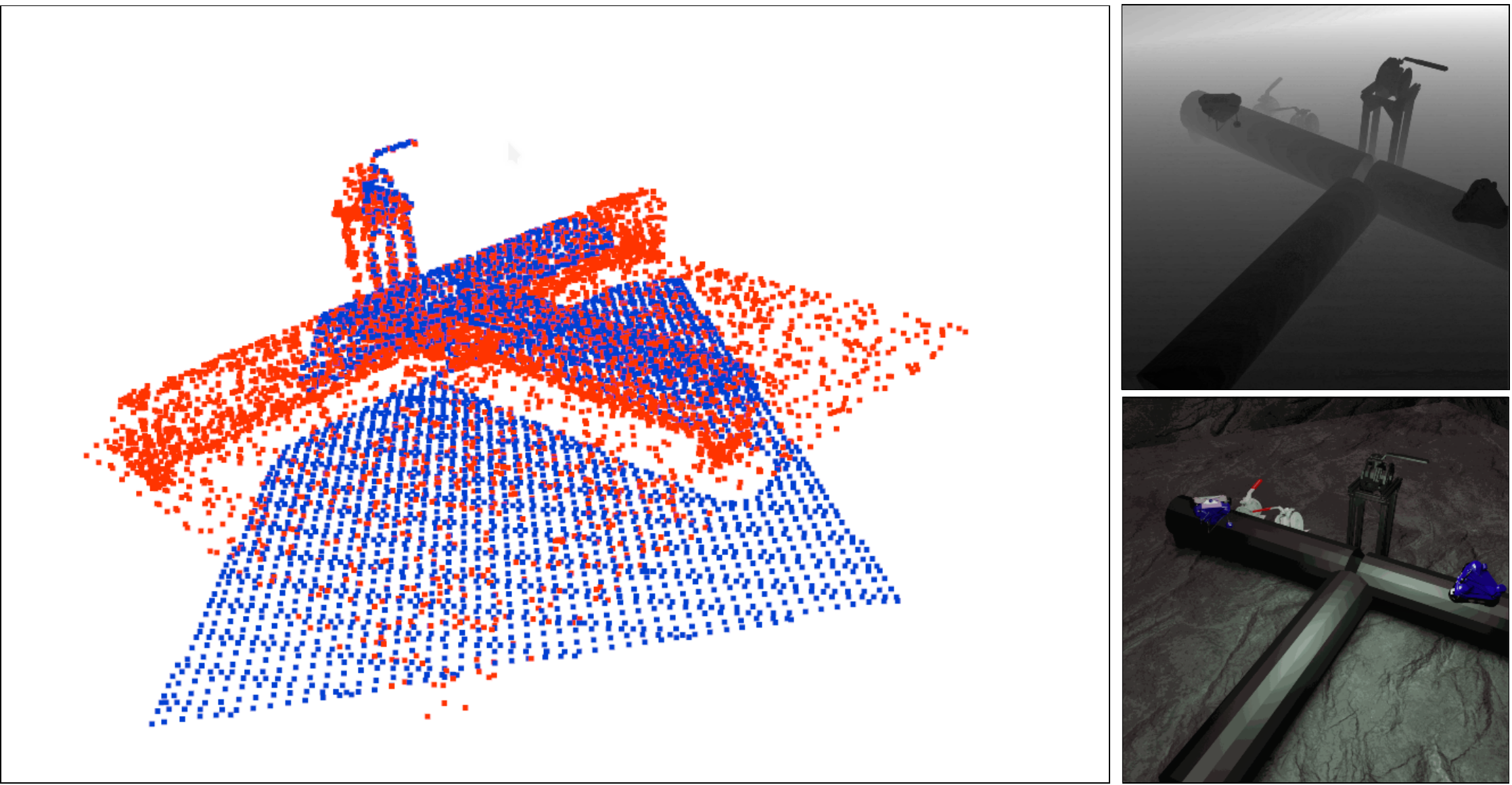}
    \caption{Illustration of the BlenderProc-based synthetic data generation pipeline. {Left:} Alignment between the digital twin point cloud (red) and the current sensor measurement (blue). {Right:} Rendering of corresponding RGB and depth images used to generate point clouds for training and evaluation.}
    \label{fig:appendix:1}
\end{figure}

For the loss function, let $\vp_i\in\mP$ and $\vq_i\in\mQ$ be matched points, and let $w_i\in(0,1)$ be the predicted inlier weight for the correspondence $(i,j)$. The training objective is a weighted rigid transformation loss:
\begin{equation}
\mathcal{L} = \frac{1}{\sum_i w_i} \sum_i w_i \, \big\| \hat{\mR}\vp_i + \hat{\vt} - \vq_i \big\|_2^2.
\end{equation}

\textbf{Testing pipeline.} During inference, the same preprocessing and correspondence generation steps are applied. The network predicts rotation and translation. To improve registration accuracy, we optionally refine the transformation using Open3D’s point-to-point ICP initialized with $\hat{\mT}$ and a distance threshold of $2\nu$. The final output includes the estimated transformation and aligned point clouds. Uncertainty estimates are inferred for both translation and Lie algebra coordinates.

\textbf{Data collection.} We generate synthetic RGB-D scenes using BlenderProc \citep{denninger2023blenderproc2} to create training and evaluation data in BOP format. Each scene contains a single object placed inside a simple room constructed from five planes (floor and four walls). The object mesh (OBJ or PLY) is loaded and scaled to meters, assigned vertex-color materials, and annotated with BOP metadata for compatibility with downstream tools. Room planes are configured as rigid bodies with high friction to prevent object penetration, and optional CC textures can be applied to the walls for additional appearance variation.

Object placement is sampled on the floor plane using BlenderProc’s surface sampling utilities. The object translation is constrained to lie above the floor with heights uniformly sampled in \([1,4]\) m, and the object rotation is sampled uniformly in yaw while pitch and roll are fixed. For each scene, camera poses are generated in two stages: first, a manually tuned keyframe pose is selected from a set of four predefined views; second, additional camera poses are produced by applying small perturbations to the keyframe. Perturbations include uniform translation noise in each axis and rotation noise applied in the Lie algebra space. Specifically, the initial rotation matrix is converted to its Lie algebra vector, noise is added, and the perturbed rotation matrix is recovered via the exponential map. A BVH-based obstacle check ensures that the camera has no geometry closer than 0.7 m, and only valid views are kept.

Lighting is randomized using a ceiling emissive plane and a point light sampled on a spherical shell, both with randomized color and intensity. Object materials are also randomized by sampling roughness and specular values; for T-LESS and ITODD, the base color is sampled uniformly in a gray range to match dataset characteristics. Rendering is performed with disabled antialiasing for depth and 50 samples for color. RGB and depth images are written in BOP format with a depth scale of 0.1 and JPEG color output. The procedure is repeated for \texttt{--num\_scenes} scenes and \texttt{--num\_cam} views per scene.

\begin{table}[t]
\centering
\caption{Synthetic data generation parameters.}
\label{tab:synthetic_params}
\begin{tabular}{l|c}
\toprule
Parameter & Value / Range \\
\midrule
Number of scenes per run & 200 (default) \\
Views per scene & 10 (default) \\
Object height above floor & $[1,4]$ m \\
Object rotation & yaw $\sim$ Uniform$[0,2\pi]$ \\
Camera distance from object & $\sim$ shell $[1.0,1.5]$ m \\
Camera obstacle clearance & $\geq 0.7$ m \\
Rotation perturbation (Lie algebra) & $\pm$ [0.1, 0.1, 0.1] rad \\
Translation perturbation & $\pm$ [0.1, 0.1, 0.1] m \\
Depth scale & 0.1 \\
Color samples & 50 \\
\bottomrule
\end{tabular}
\end{table}

\subsection{Uncertainty Estimation: Algorithmic Details}

In the main paper, we claimed efficient computations that address scalability issues of GPs, gating function based on partitioned approach to point cloud registration, and input dependent aleatoric uncertainty estimation. Below, we describe algorithmic details on the aforementioned points step by step.

\textbf{Efficient computations.} We estimate predictive uncertainty using Bayesian regression models that admit both weight-space and function-space formulations. This dual interpretation provides a principled foundation for selecting computationally efficient uncertainty estimation methods depending on the dimensionality of the model parameters and the available data.

To avoid notational ambiguity with the 6D pose output $\vy\in\mathbb{R}^6$ used in the registration task, we denote the generic scalar regression target in the following derivation by $z\in\mathbb{R}$. We first consider a linear-in-parameters model of the form
\begin{equation}
z = \phi(\vx)^\top \vw + \varepsilon, \quad \varepsilon \sim \mathcal{N}(0,\sigma_n^2),
\end{equation}
where $\phi(\vx)\in\mathbb{R}^P$ denotes a feature representation, $\vw\in\mathbb{R}^P$ is a weight vector, and $\sigma_n^2$ is the observation noise variance. A Gaussian prior $\vw\sim\mathcal{N}(\mathbf{0},\boldsymbol{\Sigma}_0)$ is assumed. Given a training dataset $\mathcal{D}=\{(\vx_i,z_i)\}_{i=1}^N$ with the following matrix:
\begin{equation}
\boldsymbol{\Phi}=[\phi(\vx_1)^\top;\dots;\phi(\vx_N)^\top],
\end{equation}
the posterior over the weights is Gaussian,
\begin{equation}
p(\vw\mid\mathcal{D})=\mathcal{N}(\boldsymbol{\mu}_{\vw},\boldsymbol{\Sigma}_{\vw}),
\end{equation}
with
\begin{align}
\boldsymbol{\Sigma}_{\vw} &= \Big(\boldsymbol{\Sigma}_0^{-1}+\frac{1}{\sigma_n^2}\boldsymbol{\Phi}^\top\boldsymbol{\Phi}\Big)^{-1}, \\
\boldsymbol{\mu}_{\vw} &= \frac{1}{\sigma_n^2}\boldsymbol{\Sigma}_{\vw}\boldsymbol{\Phi}^\top \mathbf{z}.
\end{align}
For a test input $\vx_*$, the predictive variance is given by
\begin{equation}
\sigma_*^2=\phi(\vx_*)^\top\boldsymbol{\Sigma}_{\vw}\phi(\vx_*)+\sigma_n^2,
\end{equation}
which quantifies both epistemic and aleatoric uncertainty. The computational complexity of this formulation is dominated by operations on $P\times P$ matrices, where $P$ is the number of model parameters. We note that which model parameter to account for, is a design choice. For example, one may choose to infer the probability distribution over only few last layers. Furthermore, note that formulation here is provided per output but these computations can be made parallel. Our criteria uses trace operator that only need variance (hence the total variance) and therefore in practice, we can ignore correlations between different output variables. If such correlations are important, one may use multi-GP formulation. 

To extend this framework to nonlinear models, we adopt a local linearization of a neural network $f_{\boldsymbol{\theta}}(\vx)$ around a reference parameter vector $\boldsymbol{\theta}_0$. Using a first-order Taylor expansion,
\begin{equation}
f_{\boldsymbol{\theta}}(\vx)\approx f_{\boldsymbol{\theta}_0}(\vx)+\mJ_f(\vx)(\boldsymbol{\theta}-\boldsymbol{\theta}_0),
\end{equation}
where $\mJ_f(\vx)=\frac{\partial f_{\boldsymbol{\theta}}(\vx)}{\partial \boldsymbol{\theta}}\big|_{\boldsymbol{\theta}_0}$ denotes the Jacobian of the network output with respect to the parameters. Defining
\begin{equation}
\phi(\vx)=\mJ_f(\vx)^\top, \quad \vw=\boldsymbol{\theta}-\boldsymbol{\theta}_0,
\end{equation}
the network is locally approximated by a generalized linear model. By placing a Gaussian prior on $\vw$, the resulting model is equivalent to Bayesian linear regression.

In practice, modeling all network parameters in a Bayesian manner is often computationally prohibitive due to the high dimensionality of modern deep networks. Instead, we adopt a partial Bayesian approach in which only the parameters of the last few layers are treated probabilistically, while the preceding layers are trained deterministically and serve as a feature extractor. Let $\mathbf{h}(\vx)$ denote the deterministic feature representation produced by the frozen layers. The model output can then be written as
\begin{equation}
z = {\psi}(\mathbf{h}(\vx))^\top \vw + \varepsilon,
\end{equation}
where ${\psi}(\mathbf{h}(\vx))$ represents the activations of the final Bayesian layers and $\vw$ their associated parameters.

An equivalent interpretation is obtained in function space by modeling $f(\cdot)$ as a Gaussian process with kernel
\begin{equation}
k(\vx,\vx')=\phi(\vx)^\top\boldsymbol{\Sigma}_0\phi(\vx'),
\end{equation}
which corresponds to a linear or neural tangent kernel. While this formulation avoids explicit representation of weight covariances, it requires inversion of an $N\times N$ kernel matrix and scales cubically with the number of training samples in the exact setting. By restricting Bayesian inference to the final layers and adopting the weight-space formulation, the proposed approach achieves favorable scalability and enables real-time uncertainty estimation in online. Thus, if data set size is smaller than weight size, one may use GP formulation. Otherwise, we can choose weight space covariance estimation \citep{lee2022trust}.

\textbf{Gating function.} We adopt a mixture-of-experts (MoE) framework in which the input space is partitioned based on the physical task structure. In contrast to data-driven clustering approaches, the proposed method exploits prior knowledge from the digital twin and the task-level state machine to construct physically meaningful regimes.

The manipulation process is organized as a finite-state machine with four operational modes: (i) grasping a valve, (ii) turning the valve, (iii) picking up a cage, and (iv) placing the cage onto a pipe. Each mode is associated with characteristic robot configurations, end-effector poses, and stabilized sensor viewpoints. As the suspended manipulator and base platform regulate their motion during task execution, the resulting sensor perspectives exhibit limited variability within each mode.

For each mode $m\in\{1,\dots,M\}$, a reference point cloud $\mQ_m$ is extracted from the digital twin, corresponding to the expected workspace geometry and viewing configuration. Online sensor measurements $\mP$ are registered only against the reference cloud associated with the active mode. The gating function is defined as
\begin{equation}
g_m(\vx) = \mathbb{I}(s = m),
\end{equation}
where $s$ denotes the current task state and $\mathbb{I}(\cdot)$ is the indicator function. This hard gating strategy assigns each input deterministically to its corresponding expert and incurs negligible computational overhead.

Each expert is trained to perform registration within its restricted geometric and perceptual regime. During inference, the predictive distribution is given by
\begin{equation}
p(\mathbf{y} \mid \vx) = p_m(\mathbf{y} \mid \vx), \quad \text{if } s = m,
\end{equation}
where $p_m$ denotes the model associated with the $m$-th expert.

\textbf{Input-dependent aleatoric uncertainty.} In addition to epistemic uncertainty arising from limited training data, point cloud registration is affected by input-dependent noise caused by occlusions, partial overlap, sensor artifacts, and correspondence ambiguity. To account for these effects, we employ a variance correction procedure based on the NTK formulation.

Under the NTK approximation, the network is locally linearized around a reference parameter vector, yielding a Bayesian linear model in the tangent feature space. The predictive variance for a test input $\vx_*$ can be written as
\begin{equation}
\mathrm{Var}[z_* \mid \vx_*]
=
\phi(\vx_*)^\top \boldsymbol{\Sigma}_{\vw} \phi(\vx_*) + \sigma_n^2,
\end{equation}
where the first term captures epistemic uncertainty and $\sigma_n^2$ represents observation noise.

Specifically, a correction term $c(\vx,\vx)$ is added to the kernel diagonal, resulting in the modified predictive variance
\begin{equation}
\mathrm{Var}[z_* \mid \vx_*]
=
\phi(\vx_*)^\top \boldsymbol{\Sigma}_{\vw} \phi(\vx_*) + \hat{\sigma}_\varepsilon^2(\vx_*),
\end{equation}
with
\begin{equation}
\hat{\sigma}_\varepsilon^2(\vx) = \sigma_n^2 + c(\vx,\vx).
\end{equation}

This formulation can be interpreted as introducing heteroscedastic observation noise. When applied to the 6D pose output $\vy\in\mathbb{R}^6$, we use the same mechanism dimension-wise to obtain an input-dependent diagonal covariance $\boldsymbol{\Sigma}(\vx)$ in $\mathfrak{se}(3)$. We note that we applied the ideas of \citet{zhu2025scalable} in NTK settings. More details on this technique can be found in \citet{zhu2025scalable}.

\section{System Architecture}
\label{sec:system_architecture}

\begin{figure}
    \centering
    \includegraphics[width=0.495\textwidth]{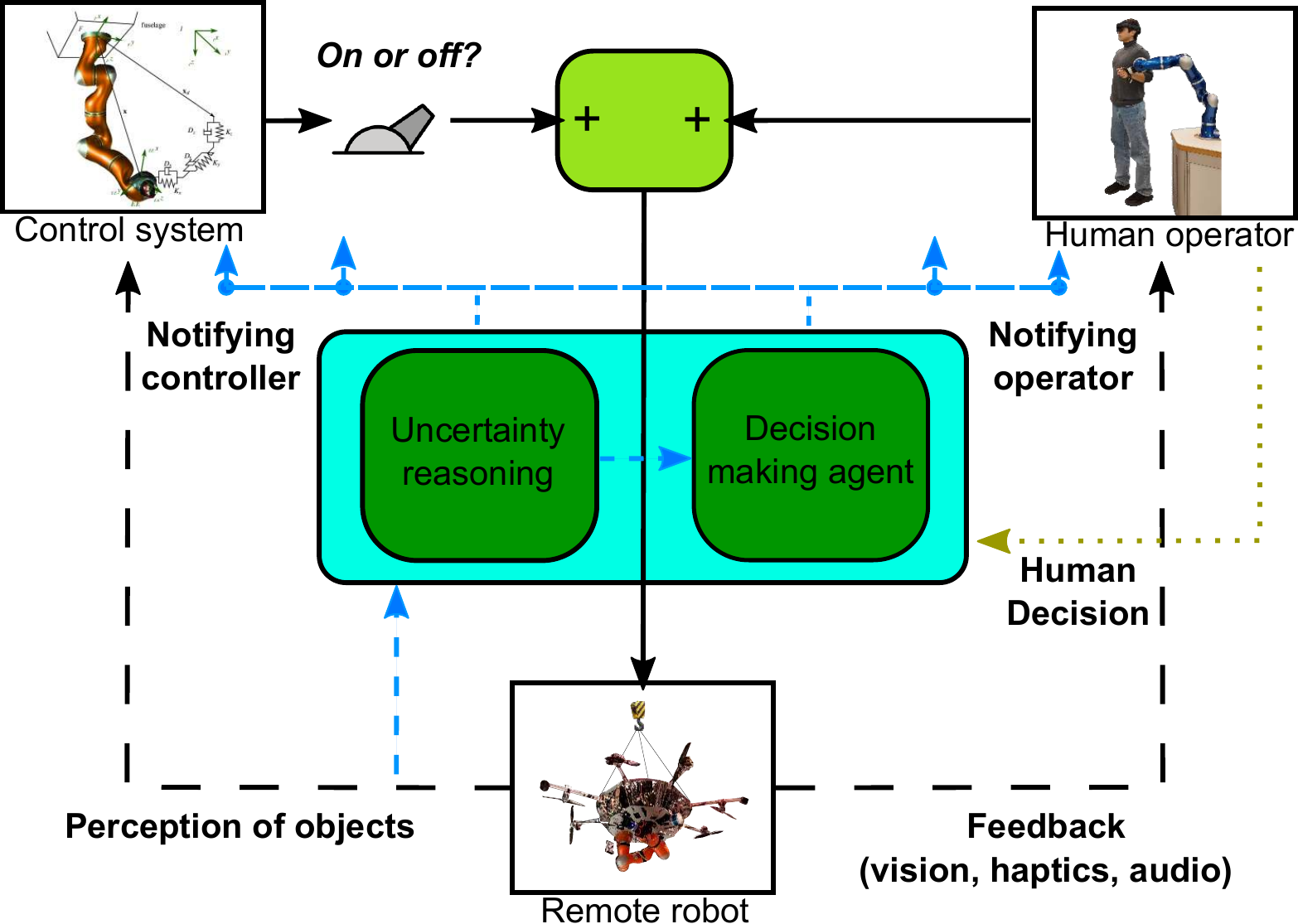}
    \caption{System architecture of the proposed perception shared autonomy.}
    \label{fig:appendix:2}
\end{figure}

Fig. \ref{fig:appendix:2} depicts the system architecture of our perception shared autonomy. The architecture consists of three main components: the remote robot, the human operator, and the robot’s autonomy, referred to as the control system. Inputs from the human operator and the autonomous controller are combined to generate a six-dimensional wrench that is applied to the torque-controlled robot. The remote robot provides multiple feedback signals, including camera images, haptic information, and audio text, which are visualized for the operator using a mixed-reality interface on the HoloLens. In addition, the robot’s perception module supplies environmental information to the control system for generating control commands. The uncertainty associated with the robot’s perception is estimated and utilized within a decision-making block. This uncertainty signal is then communicated to the control system, which enables or disables autonomous assistance in a binary manner, following a blended shared-autonomy strategy. A binary signal is adopted for simplicity and robustness. Furthermore, the estimated uncertainty is conveyed to the human operator through visual and auditory cues. Based on this information, the operator can override the system’s decision under uncertain conditions and manually enable or disable the autonomy.

\begin{figure}
    \centering
    \includegraphics[width=0.425\textwidth]{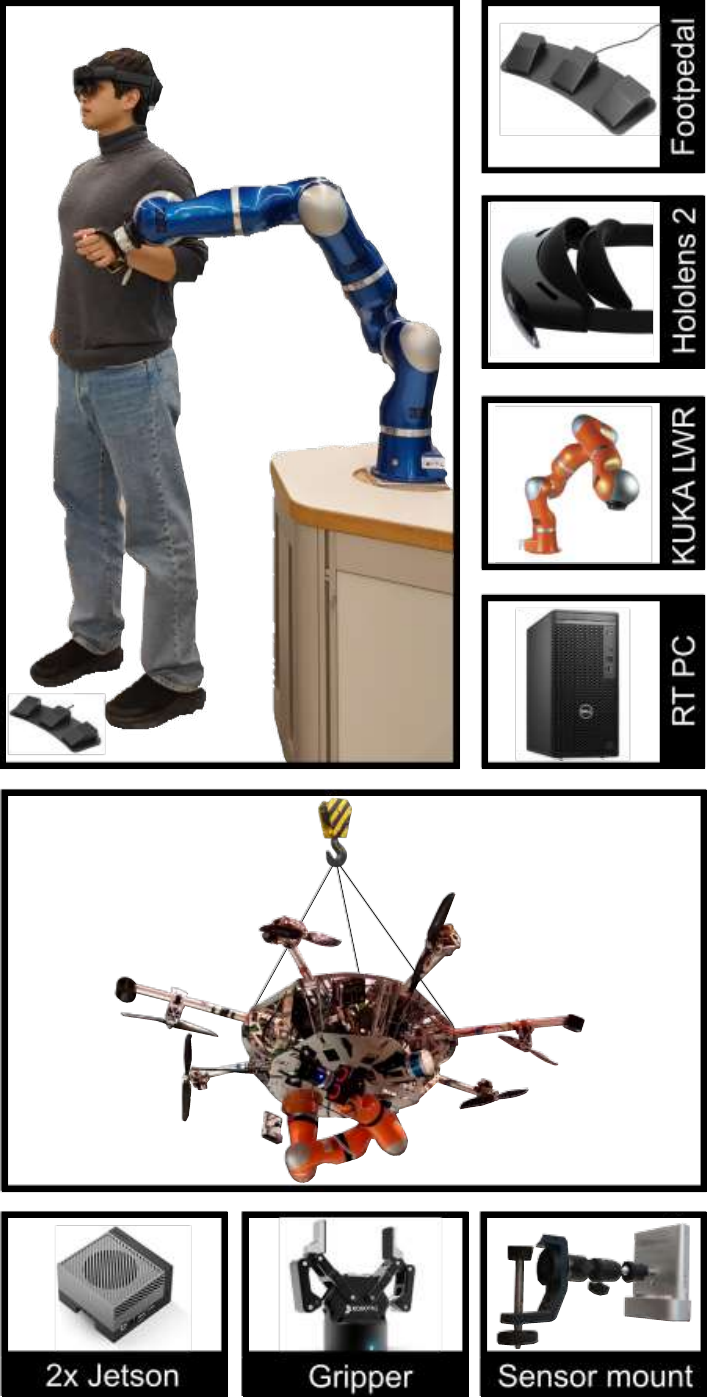}
    \caption{System components of SPIRIT. Relevant updates to the hardware are highlighted. In comparison to \citet{lee2023virtual}, new human machine interface, compute infrastructures, gripper and sensor mounts have been introduced to the overall system.}
    \label{fig:spirit:hardware}
\end{figure}

The robot hardware comprises a modular aerial manipulation system consisting of a carrier, a cable-suspended platform, and a seven-degree-of-freedom industrial robotic arm equipped with a Robotiq gripper for reliable grasping and object manipulation. The carrier, implemented using a crane in this work, transports the system to the target location and provides safety, robustness, and versatility for industrial applications, while the suspended platform, connected via cables, actively damps disturbances induced by carrier motion, environmental effects, and manipulator dynamics using eight propellers and three winches. By supporting the system’s weight through the carrier, the cable-suspension concept reduces energy consumption and enables a compact design suitable for operation in confined spaces, and the suspension cables can additionally supply power for extended operation. For teleoperation, a KUKA LBR robot is employed as a haptic interface and connected to a real-time PC to ensure low-latency and stable bilateral control, while operator feedback and system information are presented through a mixed-reality interface using the Microsoft HoloLens for immersive visualization. Advanced control strategies, including whole-body teleoperation and adaptive shared control, are implemented using passivity-based methods to guarantee stability under communication delays and network disturbances. The system integrates multiple sensors for state estimation and environmental perception, including joint torque and position sensors, RGB and RGB-D cameras, and a stereo camera mounted near the tool center point for close-range perception and depth sensing, as well as a 3D vision sensor with visual–inertial SLAM capabilities and a LiDAR sensor for navigation and object pose estimation. All perception algorithms are executed onboard using two NVIDIA Jetson Orin modules, and the sensor and computing architecture is designed to preserve close-range perception performance while enabling robust real-time processing for autonomous and shared-control operation. See Fig.~\ref{fig:spirit:hardware} for more details.

\section{User Study Details}
\label{sec:user_study}

We provide detailed information regarding the participant demographics and situational awareness questionnaire.

\textbf{Participant demographics.} To ensure a representative sample and identify potential biases related to prior technical expertise, we collected the following demographic and background information for each participant ($N = 15$):
\begin{itemize}
    \item \textit{General Information:} Anonymous Identifier, Age, Gender, and Handedness.
    \item \textit{Professional Background:} Current profession.
    \item \textit{Domain Experience:} Self-reported experience level (measured on a 5-point Likert scale) regarding:
    \begin{enumerate}
        \item Robot teleoperation for manipulation.
        \item Mixed reality (MR) interfaces.
    \end{enumerate}
\end{itemize}

\textbf{Situational awareness questionnaire.} In additional to standardized subjective metrics such as NASA Task Load Index (NASA-TLX) and System Usability Scale (SUS), we also performed quantitative analysis. To evaluate the impact of haptic feedback and MR visualization on the user's perception of the system state and uncertainty of the robot, we developed a Situational Awareness (SA) questionnaire. Each item was rated on a 8-point Likert scale, ranging from \textit{1 (Does not apply at all)} to \textit{8 (Completely applies)}.
\begin{enumerate}
    \item I felt to be aware of the positions and actions of the robots.
    \item I felt capable of anticipating the robot’s actions.
    \item I felt to be aware of the degree of control I had.
    \item I felt that haptic feedback helped in situational awareness of the robot (authority shifting).
    \item I felt that mixed reality helped in situational awareness of the robot (authority shifting)
\end{enumerate}
We reported mean and standard deviation in our paper.

\begin{figure}
    \centering
    \includegraphics[width=0.4\textwidth]{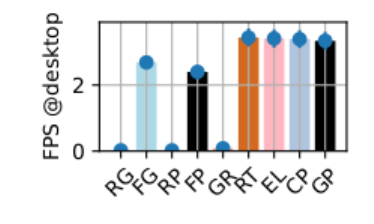}
    \caption{Runtime analysis on the standard desktop setup with NVIDIA GPU 1080Ti.}
    \label{fig:runtime}
\end{figure}

\begin{figure*}
    \centering
    \includegraphics[width=0.875\textwidth]{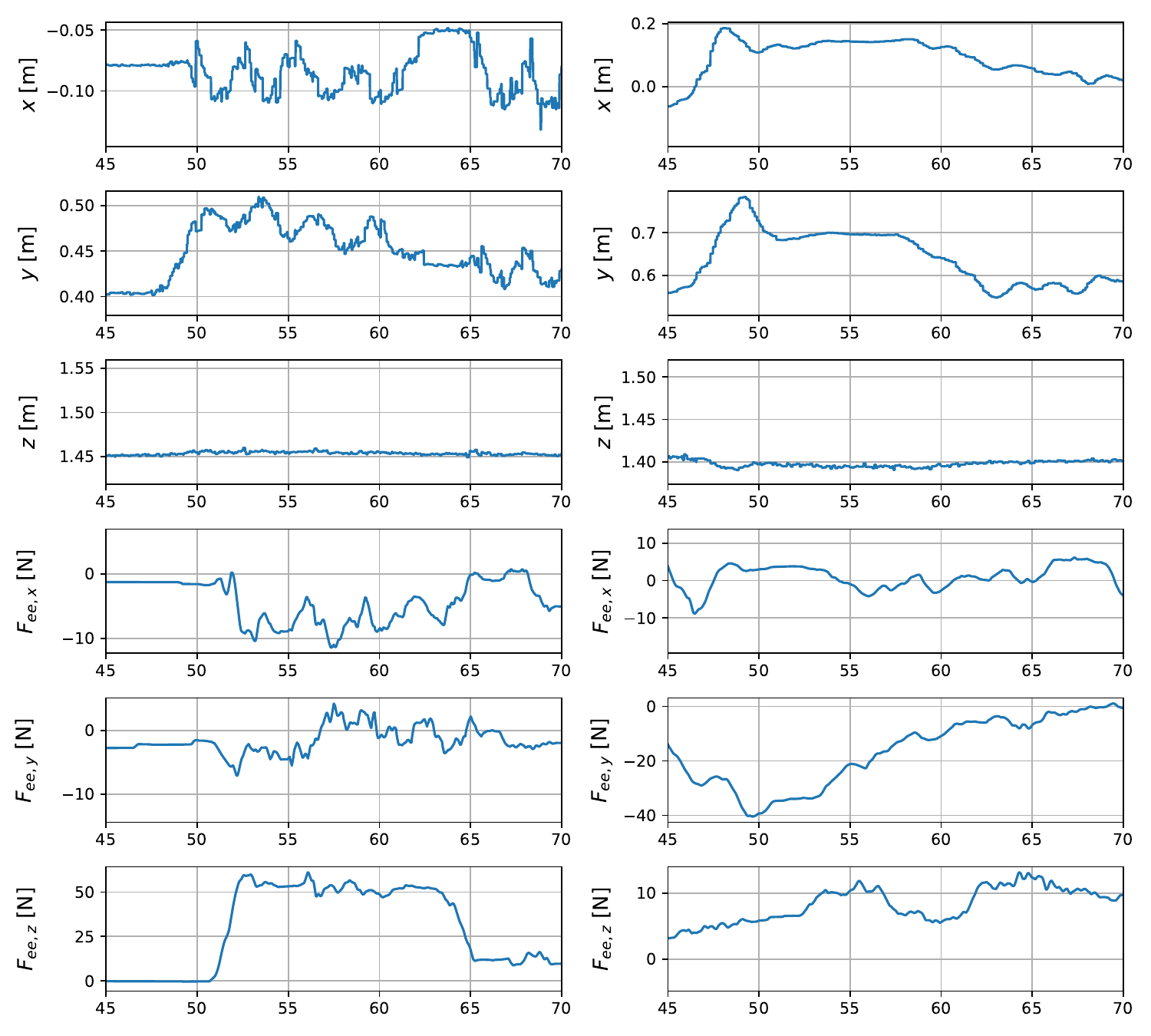}
    \caption{Additional plots from the industrial scenario. The movement of the base platform shows that the conceived experiments did not ignore floating base effects.}
    \label{fig:appendix:rebuttal}
\end{figure*}

\begin{figure*}
    \centering
    \includegraphics[width=0.975\textwidth]{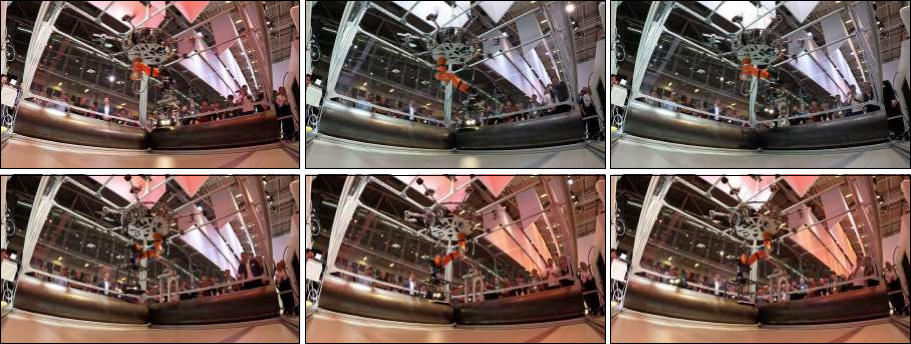}
    \caption{Demonstration of crawler deployment and retrieval at an industrial exhibition. The sequence is shown from left to right and top to bottom. The robot grasps the deployment cage, lifts and positions it, performs a pressing operation to release the crawler, and subsequently retrieves and secures the crawler after inspection. The images illustrate the full manipulation cycle under realistic operating conditions.}
    \label{fig:demo1}
\end{figure*}

\begin{figure}
    \centering
    \includegraphics[width=0.475\textwidth]{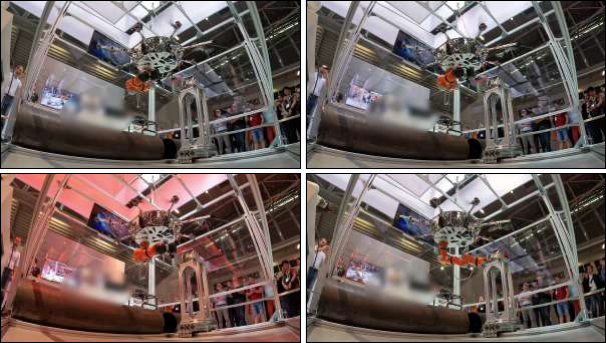}
    \caption{Demonstration of valve closing in an industrial inspection scenario. The sequence is shown from left to right and top to bottom. The robot approaches and grasps the valve handle and executes a controlled turning motion to close the valve, highlighting precise manipulation in a constrained industrial environment.}
    \label{fig:demo2}
\end{figure}

\section{Additional Details and Results}
\label{sec:additional_details_results}

\subsection{Baseline Implementations}

All baselines are implemented within a unified evaluation framework based on the GP3D codebase. For each sequence, depth images are converted to point clouds using camera intrinsics and voxel-downsampled at a fixed resolution. The first frame initializes a reference point cloud, and subsequent frames are registered relative to this reference using a transformation manager that maintains ground-truth relative poses. All experiments are conducted with fixed random seeds and identical data partitions. Registration accuracy is evaluated using rotation and translation errors, runtime is measured for each inference step, and robustness is assessed by repeating evaluation on synthetically corrupted target point clouds.

\textbf{Classical baselines.} We implement RANSAC-based global registration (RG) and Fast Global Registration (FGR) using the Open3D library \citep{zhou2018open3d}, followed by point-to-point ICP refinement. For both methods, surface normals and FPFH features are extracted from voxelized point clouds and used for feature-based correspondence matching. RG estimates an initial transformation via RANSAC with distance and geometric consistency checks, while FGR performs correspondence optimization in feature space. The resulting transformations are refined using ICP. For uncertainty-related comparisons, we compute the transformation negative log-likelihood (NLL) in the Lie algebra using a fixed isotropic covariance calibrated from a set, and we report both standard and failure NLL.

\textbf{Partitioned classical baselines.} We additionally evaluate sequential variants of RG and FGR, denoted as RP and FP, in which global registration is applied frame-to-frame without ICP refinement. Each frame is registered relative to the previous reference using known temporal ordering. This setting isolates the effect of temporal partitioning and incremental registration. We report pose errors, runtime, and NLL using the same calibration and corruption protocol as for the global baselines.

\textbf{Learning-based global registration (DGR).} We implement a deep global registration baseline \citep{choy2020deep} based on learned feature correspondences and weighted Procrustes alignment. Input point clouds are voxelized and converted to sparse tensors using MinkowskiEngine. Per-point descriptors are extracted using an FCGF backbone (or optionally FPFH features) and matched via 1-NN search in feature space. An inlier prediction network assigns correspondence-wise confidence scores, which are converted to weights via a sigmoid function and thresholded. The final pose is estimated by solving a weighted Procrustes problem. If the inlier support is insufficient, the method falls back to a RANSAC-based safeguard. ICP refinement is disabled in our evaluation. We evaluate pose accuracy, runtime, and NLL using a fixed isotropic covariance.

\textbf{End-to-end regression (RT).} We evaluate an end-to-end regression baseline that directly predicts the relative pose from matched features using a sparse MLP head. After voxelization and descriptor extraction, coarse correspondences are obtained via nearest-neighbor matching. Correspondence features are processed by a MinkowskiEngine-based network that regresses rotation and translation parameters together with confidence logits. Unlike DGR, no explicit geometric solver is used, and the final pose is obtained directly from the network output. ICP refinement is disabled to isolate the learned estimator. We report pose errors, runtime, and NLL using the same fixed covariance as for DGR. We note that the difference to DGR \citep{choy2020deep} is not using weighted Procrustes solver, but directly regressing the registration results. This design choice is to test our idea on shared autonomy rather than to improve point cloud registration.

\textbf{Evidential learning.} We implement an evidential regression \citep{amini2020deep} that extends RT by explicitly modeling predictive uncertainty. The network outputs Normal-Inverse-Gamma (NIG) parameters $(\mu, v, \alpha, \beta)$ for both rotation (in Lie algebra) and translation. These parameters define a predictive mean and variance, allowing the model to represent data-dependent aleatoric uncertainty. The final pose is obtained from the predicted means, while a diagonal covariance in $\mathfrak{se}(3)$ is computed from the closed-form NIG variance. Training is performed using an evidential loss combining negative log-likelihood and evidence regularization. At inference time, we evaluate pose errors, runtime, and NLL using the predicted per-sample variance. Various hyperparameters (epochs, learning rate, and batch size) in grid setting was tried.

\textbf{Conformal prediction.} As a post-hoc uncertainty calibration baseline, we apply split conformal prediction \citep{fontana2023conformal} to deterministic model (RT). Using a held-out calibration set, we collect predicted and ground-truth poses in Lie algebra and compute empirical quantiles of absolute residuals at miscoverage level $\alpha$. These quantiles define constant, per-dimension error bounds that are converted into diagonal covariance estimates. The resulting variance is optionally scaled using the same variance optimization procedure as in other baselines. During testing, predictions are refined using ICP and evaluated using pose errors, runtime, and NLL with conformal uncertainty.

\textbf{NTK-based aleatoric calibration.} For baselines that do not explicitly predict uncertainty (RG, FGR, RP, FP, DGR, RT), we additionally evaluate an NTK-based variance calibration procedure. Using the batch of sample in each sequence, which we call calibration dataset, we optimize a global isotropic variance parameter in the Lie algebra by minimizing the transformation NLL between predicted and ground-truth poses. The resulting variance is treated as an input-dependent aleatoric proxy and used for subsequent NLL evaluation.

\subsection{Run-time Analysis}

In Fig.~\ref{fig:runtime}, we compare the runtime performance of all baseline methods. We observe that RANSAC-based approaches exhibit the highest computational cost, followed by learning-based global registration with weighted Procrustes, due to the overhead of correspondence sampling and iterative optimization. In contrast, Fast Global Registration and its partitioned variant achieve substantially lower runtimes, benefiting from efficient correspondence optimization and reduced computational complexity. The end-to-end regression model (RT), which represents a standard neural-network-based registration approach, further improves computational efficiency by avoiding explicit geometric solvers. Importantly, we find that the incorporation of uncertainty estimation mechanisms, including NTK-based calibration, evidential regression, and conformal prediction, does not introduce significant additional latency, as these methods rely on lightweight post-processing or closed-form computations. Overall, the results indicate that uncertainty-aware registration can be achieved without compromising real-time performance. For faster run-rate, if required by controller, we can also combine online SLAM estimates of robot's state.

\subsection{Industrial Scenaiors}

For our experiments, we are not ignoring key constraints unique to aerial platforms. Fig. \ref{fig:appendix:rebuttal} further depicts the robot's position and interaction forces during lifting and placing of heavy objects as well as valve closing. We observe that large displacements are found which SPIRIT is able to cope with. This is thanks to previous work on control and hardware, and in this paper, we are able to advance the autonomy and perception modules of our aerial manipulation research.

\subsection{Industrial Exhibition}

SPIRIT was demonstrated at major international industrial fairs in robotics and automation, highlighting its applicability to real-world inspection and maintenance scenarios. Fig.~\ref{fig:demo1} illustrates the crawler deployment and retrieval tasks, while Fig.~\ref{fig:demo2} presents the valve-closing demonstration. In these setups, the inspection robot crawler, pipeline structure, and valve assemblies are clearly visible, emphasizing the realism of the experimental environment. The demonstrations attracted substantial attention from both industry and media. As shown in Fig.~\ref{fig:demo1}, the robot sequentially grasps, lifts, and positions the deployment cage, followed by a pressing operation to release the crawler for inspection. During retrieval, the robot performs the inverse procedure to secure the crawler. In Fig.~\ref{fig:demo2}, the robot grasps the valve handle and executes a turning motion to close the valve, demonstrating precise manipulation under constrained industrial conditions. Together, these experiments showcase the practical relevance of SPIRIT for robotic inspection and maintenance applications.

%% file: bibliography.bib
@article{thrun2000probabilistic,
  title={Probabilistic algorithms and the interactive museum tour-guide robot minerva},
  author={Thrun, Sebastian and Beetz, Michael and Bennewitz, Maren and Burgard, Wolfram and Cremers, Armin B and Dellaert, Frank and Fox, Dieter and Haehnel, Dirk and Rosenberg, Chuck and Roy, Nicholas and others},
  journal={International Journal of Robotics Research (IJRR)},
  volume={19},
  number={11},
  pages={972--999},
  year={2000}
}

@article{sunderhauf2018limits,
  title={The limits and potentials of deep learning for robotics},
  author={S{\"u}nderhauf, Niko and Brock, Oliver and Scheirer, Walter and Hadsell, Raia and Fox, Dieter and Leitner, J{\"u}rgen and Upcroft, Ben and Abbeel, Pieter and Burgard, Wolfram and Milford, Michael and others},
  journal={International Journal of Robotics Research (IJRR)},
  volume={37},
  number={4-5},
  pages={405--420},
  year={2018},
  publisher={SAGE Publications Sage UK: London, England}
}

@article{gunning2019xai,
  title={XAI—Explainable artificial intelligence},
  author={Gunning, David and Stefik, Mark and Choi, Jaesik and Miller, Timothy and Stumpf, Simone and Yang, Guang-Zhong},
  journal={Science robotics},
  volume={4},
  number={37},
  pages={eaay7120},
  year={2019},
  publisher={American Association for the Advancement of Science}
}

@article{billard2025roadmap,
  title={A roadmap for AI in robotics},
  author={Billard, Aude and Albu-Schaeffer, Alin and Beetz, Michael and Burgard, Wolfram and Corke, Peter and Ciocarlie, Matei and Dahiya, Ravinder and Kragic, Danica and Goldberg, Ken and Nagai, Yukie and others},
  journal={Nature Machine Intelligence},
  pages={1--7},
  year={2025},
  publisher={Nature Publishing Group UK London}
}

@article{dellaert2012factor,
  title={Factor graphs and GTSAM: A hands-on introduction},
  author={Dellaert, Frank},
  journal={Georgia Institute of Technology, Tech. Rep},
  volume={2},
  pages={4},
  year={2012}
}

@inproceedings{feng2023topology,
  title={Topology-Matching Normalizing Flows for Out-of-Distribution Detection in Robot Learning},
  author={Feng, Jianxiang and Lee, Jongseok and Geisler, Simon and G{\"u}nnemann, Stephan and Triebel, Rudolph},
  booktitle={Conference on Robot Learning (CoRL)},
  year={2023}
}

@inproceedings{choy2020deep,
  title={Deep global registration},
  author={Choy, Christopher and Dong, Wei and Koltun, Vladlen},
  booktitle={IEEE Conference on Computer Vision and Pattern Recognition (CVPR)},
  year={2020}
}

@inproceedings{coelho2021hierarchical,
  title={Hierarchical control of redundant aerial manipulators with enhanced field of view},
  author={Coelho, Andre and Sarkisov, Yuri S and Lee, Jongseok and Balachandran, Ribin and Franchi, Antonio and Kondak, Konstantin and Ott, Christian},
  booktitle={International Conference on Unmanned Aircraft Systems (ICUAS)},
  year={2021}
}

@book{barfoot2024state,
  title={State estimation for robotics},
  author={Barfoot, Timothy D},
  year={2024},
  publisher={Cambridge University Press}
}

@article{ollero2018aeroarms,
  title={The aeroarms project: Aerial robots with advanced manipulation capabilities for inspection and maintenance},
  author={Ollero, Anibal and Heredia, Guillermo and Franchi, Antonio and Antonelli, Gianluca and Kondak, Konstantin and Sanfeliu, Alberto and Viguria, Antidio and Martinez-de Dios, J Ramiro and Pierri, Francesco and Cort{\'e}s, Juan and others},
  journal={IEEE Robotics and Automation Magazine},
  year={2018}
}

@INPROCEEDINGS{Fox-RSS-19, 
    AUTHOR    = {Fabio Ramos AND Rafael Possas AND Dieter Fox}, 
    TITLE     = {BayesSim: Adaptive Domain Randomization Via Probabilistic Inference for Robotics Simulators}, 
    BOOKTITLE = {Robotics, Systems and Science (RSS)}, 
    YEAR      = {2019}, 
    MONTH     = {June}
}

@article{Agha2021NeBulaQF,
	title={NeBula: Quest for Robotic Autonomy in Challenging Environments; TEAM CoSTAR at the DARPA Subterranean Challenge},
	author={Agha, A. and Otsu, K. and Morrell, B. and Fan, D. D. and Thakker, R.and Santamaria-Navarro, A. and ... Burdick, J.},
	journal={Field Robotics},
	volume={2},
	pages={1432–1506},
	year={2022}
}

@inproceedings{roy1999coastal,
  title={Coastal navigation-mobile robot navigation with uncertainty in dynamic environments},
  author={Roy, Nicholas and Burgard, Wolfram and Fox, Dieter and Thrun, Sebastian},
  booktitle={IEEE International Conference on Robotics and Automation (ICRA)},
  year={1999}
}

@article{fox2000probabilistic,
  title={A probabilistic approach to collaborative multi-robot localization},
  author={Fox, Dieter and Burgard, Wolfram and Kruppa, Hannes and Thrun, Sebastian},
  journal={Autonomous robots},
  volume={8},
  number={3},
  pages={325--344},
  year={2000},
  publisher={Springer}
}

@article{ollero2021past,
  title={Past, present, and future of aerial robotic manipulators},
  author={Ollero, Anibal and Tognon, Marco and Suarez, Alejandro and Lee, Dongjun and Franchi, Antonio},
  journal={IEEE Transactions on Robotics (T-RO)},
  volume={38},
  number={1},
  pages={626--645},
  year={2021},
  publisher={IEEE}
}

@article{miyazaki2019long,
  title={Long-reach aerial manipulation employing wire-suspended hand with swing-suppression device},
  author={Miyazaki, Ryo and Jiang, Rui and Paul, Hannibal and Huang, Yanzhao and Shimonomura, Kazuhiro},
  journal={IEEE Robotics and Automation Letters},
  volume={4},
  number={3},
  pages={3045--3052},
  year={2019},
  publisher={IEEE}
}

@article{perozo2024teleoperation,
  title={Teleoperation of a Suspended Aerial Manipulator Using a Handheld Camera with an IMU},
  author={Perozo, Miguel Arpa and Niddam, Ethan and Durand, Sylvain and Cuvillon, Lo{\"\i}c and Gangloff, Jacques},
  journal={IEEE Robotics and Automation Letters (RA-L)},
  year={2024}
}

@article{xiangdong2020asset,
  title={Asset management of oil and gas pipeline system Based on Digital Twin},
  author={Xiangdong, Xue and Bo, Li and Jiannan, Gai},
  journal={IFAC-PapersOnLine},
  year={2020}
}

@ARTICLE{lee2024,
  author={Lee, Jongseok and Balachandran, Ribin and Kondak, Konstantin and Coelho, Andre and Stefano, Marco de and Humt, Matthias and Feng, Jianxiang and Asfour, Tamim and Triebel, Rudolph},
  journal={IEEE Transactions on Field Robotics (T-FR)}, 
  title={Introspective Perception for Long-Term Aerial Telemanipulation With Virtual Reality}, 
  year={2024},
  volume={1},
  number={},
  pages={360-393}
}

@article{lee2023virtual,
  title={Virtual Reality via Object Pose Estimation and Active Learning: Realizing Telepresence Robots with Aerial Manipulation Capabilities},
  author={Lee, Jongseok and Radhakrishna Balachandran, Ribin and Kondak, Konstantin and Coelho, Andre and De Stefano, Marco and Humt, Matthias and Feng, Jianxiang and Asfour, Tamim and Triebel, Rudolph},
  journal={Field Robotics},
  volume={3},
  pages={323--367},
  year={2023},
  publisher={Field Robotics Publication Society}
}

@inproceedings{wang2016apriltag,
  title={AprilTag 2: Efficient and robust fiducial detection},
  author={Wang, John and Olson, Edwin},
  booktitle={IEEE International Conference on Intelligent Robots and Systems (IROS)},
  pages={4193--4198},
  year={2016}
}

@article{zhou2018open3d,
  title={Open3D: A modern library for 3D data processing},
  author={Zhou, Qian-Yi and Park, Jaesik and Koltun, Vladlen},
  journal={arXiv preprint arXiv:1801.09847},
  year={2018}
}

@article{amini2020deep,
  title={Deep evidential regression},
  author={Amini, Alexander and Schwarting, Wilko and Soleimany, Ava and Rus, Daniela},
  journal={Neural Information Processing Systems (NeurIPS)},
  volume={33},
  pages={14927--14937},
  year={2020}
}

@article{fontana2023conformal,
  title={Conformal prediction: a unified review of theory and new challenges},
  author={Fontana, Matteo and Zeni, Gianluca and Vantini, Simone},
  journal={Bernoulli},
  volume={29},
  number={1},
  pages={1--23},
  year={2023},
  publisher={Bernoulli Society for Mathematical Statistics and Probability}
}

@inproceedings{aarno2005adaptive,
  title={Adaptive virtual fixtures for machine-assisted teleoperation tasks},
  author={Aarno, Daniel and Ekvall, Staffan and Kragic, Danica},
  booktitle={IEEE International Conference on Robotics and Automation (ICRA)},
  pages={1139--1144},
  year={2005}
}

@article{saeidi2018incorporating,
  title={Incorporating trust and self-confidence analysis in the guidance and control of (semi) autonomous mobile robotic systems},
  author={Saeidi, Hamed and Wang, Yue},
  journal={IEEE Robotics and Automation Letters (RA-L)},
  volume={4},
  number={2},
  pages={239--246},
  year={2018}
}

@inproceedings{palmieri2024perception,
  title={Perception-Driven Shared Control Architecture for Agricultural Robots Performing Harvesting Tasks},
  author={Palmieri, Jozsef and Di Lillo, Paolo and Sanfeliu, Alberto and Marino, Alessandro},
  booktitle={IEEE International Conference on Intelligent Robots and Systems (IROS)},
  year={2024}
}

@article{raiola2018co,
  title={Co-manipulation with a library of virtual guiding fixtures},
  author={Raiola, Gennaro and Restrepo, Susana Sanchez and Chevalier, Pauline and Rodriguez-Ayerbe, Pedro and Lamy, Xavier and Tliba, Sami and Stulp, Freek},
  journal={Autonomous Robots},
  volume={42},
  pages={1037--1051},
  year={2018},
  publisher={Springer}
}

@inproceedings{balachandran2020adaptive,
  title={Adaptive authority allocation in shared control of robots using Bayesian filters},
  author={Balachandran, Ribin and Mishra, Hrishik and Cappelli, Matteo and Weber, Bernhard and Secchi, Cristian and Ott, Christian and Albu-Schaeffer, Alin},
  booktitle={IEEE International Conference on Robotics and Automation (ICRA)},
  pages={11298--11304},
  year={2020}
}

@article{yow2023shared,
  title={Shared Autonomy of a Robotic Manipulator for Grasping under Human Intent Uncertainty using POMDPs},
  author={Yow, J-Anne and Garg, Neha Priyadarshini and Ang, Wei Tech},
  journal={IEEE Transactions on Robotics (T-RO)},
  year={2023}
}

@INPROCEEDINGS{Zhao-RSS-24, 
    AUTHOR    = {Michelle D Zhao AND Reid Simmons AND Henny Admoni AND Andrea Bajcsy}, 
    TITLE     = {{Conformalized Teleoperation: Confidently Mapping Human Inputs to High-Dimensional Robot Actions}}, 
    BOOKTITLE = {Robotics, Systems and Science (RSS)}, 
    YEAR      = {2024},
    MONTH     = {July}
}

@article{hara2023uncertainty,
  title={Uncertainty-aware haptic shared control with humanoid robots for flexible object manipulation},
  author={Hara, Takumi and Sato, Takashi and Ogata, Tetsuya and Awano, Hiromitsu},
  journal={IEEE Robotics and Automation Letters (RA-L)},
  year={2023}
}

@inproceedings{schnaus2023learning,
  title={Learning expressive priors for generalization and uncertainty estimation in neural networks},
  author={Schnaus, Dominik and Lee, Jongseok and Cremers, Daniel and Triebel, Rudolph},
  booktitle={International Conference on Machine Learning (ICML)},
  year={2023}
}

@article{gawlikowski2023survey,
  title={A survey of uncertainty in deep neural networks},
  author={Gawlikowski, Jakob and Tassi, Cedrique Rovile Njieutcheu and Ali, Mohsin and Lee, Jongseok and Humt, Matthias and Feng, Jianxiang and Kruspe, Anna and Triebel, Rudolph and Jung, Peter and Roscher, Ribana and others},
  journal={Artificial Intelligence Review},
  volume={56},
  pages={1513--1589},
  year={2023},
  publisher={Springer}
}

@inproceedings{gal2016dropout,
 author = {Gal, Yarin and Ghahramani, Zoubin},
 title = {Dropout as a bayesian approximation: Representing model uncertainty in deep learning},
 booktitle = {International Conference on Machine Learning (ICML)},
 year = {2016}
}

@article{lakshminarayanan2017simple,
  title={Simple and scalable predictive uncertainty estimation using deep ensembles},
  author={Lakshminarayanan, Balaji and Pritzel, Alexander and Blundell, Charles},
  journal={Neural Information Processing Systems (NeurIPS)},
  volume={30},
  year={2017}
}

@article{muhlbauer2024probabilistic,
  title={A Probabilistic Approach to Multi-Modal Adaptive Virtual Fixtures},
  author={M{\"u}hlbauer, Maximilian and Hulin, Thomas and Weber, Bernhard and Calinon, Sylvain and Stulp, Freek and Albu-Sch{\"a}ffer, Alin and Silv{\'e}rio, Jo{\~a}o},
  journal={IEEE Robotics and Automation Letters (RA-L)},
  year={2024}
}

@article{jacobs1991adaptive,
  title={Adaptive mixtures of local experts},
  author={Jacobs, Robert A and Jordan, Michael I and Nowlan, Steven J and Hinton, Geoffrey E},
  journal={Neural computation},
  volume={3},
  number={1},
  pages={79--87},
  year={1991},
  publisher={MIT Press}
}

@article{zhu2025scalable,
  title={Scalable Gaussian Processes with Low-Rank Deep Kernel Decomposition},
  author={Zhu, Yunqin and Yuchi, Henry Shaowu and Xie, Yao},
  journal={arXiv preprint arXiv:2505.18526},
  year={2025}
}

@inproceedings{lee2022trust,
  title={Trust your robots! predictive uncertainty estimation of neural networks with sparse gaussian processes},
  author={Lee, Jongseok and Feng, Jianxiang and Humt, Matthias and M{\"u}ller, Marcus Gerhard and Triebel, Rudolph},
  booktitle={Conference on Robot Learning (CoRL)},
  year={2022}
}

@incollection{rasmussen2003gaussian,
  title={Gaussian processes in machine learning},
  author={Rasmussen, Carl Edward},
  booktitle={Summer school on machine learning},
  pages={63--71},
  year={2003},
  publisher={Springer}
}

@article{kong2024suspended,
  title={A suspended aerial manipulation avatar for physical interaction in unstructured environments},
  author={Kong, Fanyi and Zambella, Grazia and Monteleone, Simone and Grioli, Giorgio and Catalano, Manuel G and Bicchi, Antonio},
  journal={IEEE Access},
  volume={12},
  pages={48108--48125},
  year={2024}
}

@article{selvaggio2021autonomy,
  title={Autonomy in physical human-robot interaction: A brief survey},
  author={Selvaggio, Mario and Cognetti, Marco and Nikolaidis, Stefanos and Ivaldi, Serena and Siciliano, Bruno},
  journal={IEEE Robotics and Automation Letters (RA-L)},
  volume={6},
  number={4},
  pages={7989--7996},
  year={2021}
}

@article{lee2025human,
  title={Human-in-the-Loop Gaussian Splatting for Robotic Teleoperation},
  author={Lee, Yongseok and Kim, Hyunsu and Ji, Harim and Heo, Jinuk and Lee, Youngseon and Kang, Jiseock and Lee, Jeongseob and Lee, Dongjun},
  journal={IEEE Robotics and Automation Letters},
  volume={11},
  number={1},
  pages={105--112},
  year={2025},
  publisher={IEEE}
}

@inproceedings{lee2020estimating,
  title={Estimating model uncertainty of neural networks in sparse information form},
  author={Lee, Jongseok and Balachandran, Ribin and Sarkisov, Yuri S and De Stefano, Marco and Coelho, Andre and Shinde, Kashmira and Kim, Min Jun and Triebel, Rudolph and Kondak, Konstantin},
  booktitle={International Conference on Machine Learning (ICML)},
  year={2020}
}

@inproceedings{sarkisov2019development,
  title={Development of sam: cable-suspended aerial manipulator},
  author={Sarkisov, Yuri S and Kim, Min Jun and Bicego, Davide and Tsetserukou, Dzmitry and Ott, Christian and Franchi, Antonio and Kondak, Konstantin},
  booktitle={IEEE International Conference on Robotics and Automation (ICRA)},
  pages={5323--5329},
  year={2019}
}

@inproceedings{zurek2021situational,
  title={Situational confidence assistance for lifelong shared autonomy},
  author={Zurek, Matthew and Bobu, Andreea and Brown, Daniel S and Dragan, Anca D},
  booktitle={IEEE International Conference on Robotics and Automation (ICRA)},
  pages={2783--2789},
  year={2021}
}

@inproceedings{hoquethriftydagger,
  title={ThriftyDAgger: Budget-Aware Novelty and Risk Gating for Interactive Imitation Learning},
  author={Hoque, Ryan and Balakrishna, Ashwin and Novoseller, Ellen and Wilcox, Albert and Brown, Daniel S and Goldberg, Ken},
  year={2022},
  booktitle={Conference on Robot Learning (CoRL)}
}

@inproceedings{lee2020visual,
  title={Visual-inertial telepresence for aerial manipulation},
  author={Lee, Jongseok and Balachandran, Ribin and Sarkisov, Yuri S and De Stefano, Marco and Coelho, Andre and Shinde, Kashmira and Kim, Min Jun and Triebel, Rudolph and Kondak, Konstantin},
  booktitle={IEEE International Conference on Robotics and Automation (ICRA)},
  pages={1222--1229},
  year={2020}
}

@article{denninger2023blenderproc2,
  title={Blenderproc2: A procedural pipeline for photorealistic rendering},
  author={Denninger, Maximilian and Winkelbauer, Dominik and Sundermeyer, Martin and Boerdijk, Wout and Knauer, Markus and Strobl, Klaus H and Humt, Matthias and Triebel, Rudolph},
  journal={Journal of Open Source Software},
  year={2023}
}
